\documentclass{article}

\usepackage[preprint]{neurips_2024}

\usepackage[utf8]{inputenc} 
\usepackage[T1]{fontenc}    
\usepackage{hyperref}       
\usepackage{url}            
\usepackage{booktabs}       
\usepackage{amsfonts}       
\usepackage{nicefrac}       
\usepackage{microtype}      
\usepackage{xcolor}         

\usepackage{amsmath}             
\usepackage{amssymb}             
\usepackage{amsthm}              
\usepackage{graphicx}            
\usepackage{subcaption}          
\usepackage{float}               
\usepackage{multirow}            
\usepackage{tabularx}            
\usepackage{siunitx}             

\title{Regression Augmentation With Data-Driven Segmentation}

\author{%
  Shayan Alahyari \\
  Department of Computer Science\\
  Western University\\
  London, Ontario, Canada \\
  \texttt{salahya@uwo.ca} \\
  \And
  Shiva Mehdipour Ghobadlou \\
  Department of Statistical and Actuarial Sciences\\
  Western University\\
  London, Ontario, Canada \\
  \texttt{smehdipo@uwo.ca} \\
  \And
  Mike Domaratzki \\
  Department of Computer Science\\
  Western University\\
  London, Ontario, Canada \\
  \texttt{mdomarat@uwo.ca} \\
}

\begin{document}

\maketitle

\begin{abstract}
Imbalanced regression arises when the target distribution is skewed, causing models to focus on dense regions and struggle with underrepresented (minority) samples. Despite its relevance across many applications, few methods have been designed specifically for this challenge. Existing approaches often rely on fixed, ad hoc thresholds to label samples as rare or common, overlooking the continuous complexity of the joint feature-target space and fail to represent the true underlying rare regions. To address these limitations, we propose a fully data-driven GAN-based augmentation framework that uses Mahalanobis-Gaussian Mixture Modeling (GMM) to automatically identify minority samples and employs deterministic nearest-neighbour matching to enrich sparse regions. Rather than preset thresholds, our method lets the data determine which observations are truly rare. Evaluation on 32 benchmark imbalanced regression datasets demonstrates that our approach consistently outperforms state-of-the-art data augmentation methods.
\end{abstract}


\section{Introduction}

\begin{figure*}[t]
  \centering
  \includegraphics[width=0.8\textwidth]{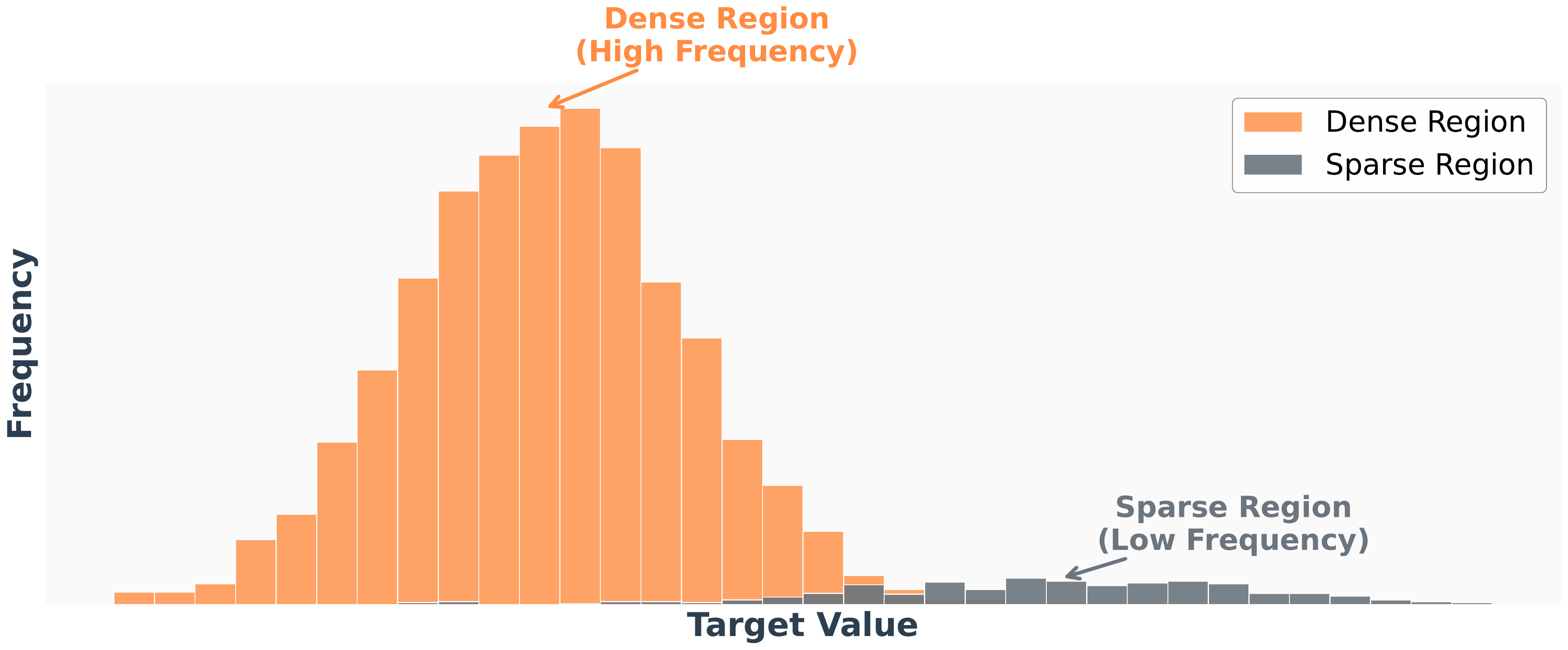}
  \caption{Distribution of target values in an imbalanced regression dataset.}
  \label{fig:imbalance_problem}
\end{figure*}

Imbalanced regression occurs when a continuous target’s distribution is skewed, leaving extreme or infrequent values underrepresented and impairing model performance \citep{krawczyk2016}. Standard regressors optimise a global error metric and thus focus on densely sampled regions, neglecting rare values and producing large errors on extremes \citep{branco2016,torgo2013}. Yet accurate prediction of these infrequent outcomes is vital in many real-world applications \citep{chawla2004,branco2016,torgo2013}.

In contrast, imbalanced classification has been studied much more than imbalanced regression. This situation arises when some classes have far fewer examples, causing models to favour majority classes and misdetect minority patterns, yielding high overall accuracy but poor rare-class detection \citep{haixiang2017,johnson2019}. Domains like fraud detection, medical diagnostics and fault detection depend on recognising these rare classes, so techniques such as oversampling, undersampling, cost-sensitive learning and ensemble methods are essential \citep{he2009,chawla2002,buda2018,liu2009}.

Figure~\ref{fig:imbalance_problem} demonstrates the distribution of target values in an imbalanced regression dataset. The dense region contains the majority of samples with high frequency, while the sparse region has significantly fewer samples, making it challenging for machine learning models to accurately predict values in the underrepresented areas.

Imbalanced regression impacts many critical real-world applications where accurately predicting rare but high-impact events in the tails of skewed distributions is paramount. For instance, in precision agriculture, vineyard yield datasets feature extreme low- and high-yield blocks that are poorly represented, prompting conditional UNet–ConvLSTM architectures with zonal weighting to address these extremes \citep{kamangir2024large}. In chemical synthesis planning, high-yield reactions are rarer yet more valuable, yet existing prediction models overlook this imbalance and underperform on critical high-yield bins \citep{ma2024revisiting}. In meteorology, rare rapid intensification of wind speed is underforecast by deep ConvLSTM models unless reweighted for the right-tail distribution \citep{scheepens2023adapting}. In materials engineering, autoencoder-based models with adaptive kernel density weighting correct underrepresentation of extreme steel plate yield strengths \citep{zhang2025irmae}. These instances highlight the limitations of traditional regression methods and emphasize the necessity for dedicated imbalanced-regression techniques.

Most oversampling methods impose domain knowledge and ad hoc thresholds to mark rarity, oversimplifying the continuous target structure and often degrading generalization. We propose a fully data-driven pipeline, requiring no arbitrary thresholds, that flags minority points via Mahalanobis distance with Gaussian mixture modelling and then uses a Wasserstein GAN to generate realistic synthetic samples, improving both rare-value accuracy and overall performance. To our knowledge, this is the first approach to combine Mahalanobis-GMM detection with GAN-based generation for imbalanced regression, eliminating the need for manual threshold selection while preserving the statistical properties of minority samples. Our method is specifically designed for tabular data, addressing a critical need in structured data applications.

\section{Related work}

\subsection{Class imbalance}
We begin with a brief survey of relevant class imbalance techniques from classification. \citet{chawla2002} introduced the first data-level class imbalance handling approach, SMOTE, which generated synthetic minority samples \(\tilde{\mathbf{x}}\) by interpolating between a minority instance \(\mathbf{x}_i\) and one of its \(k\) nearest neighbours \(\mathbf{x}_{i,\mathrm{nn}}\) according to
\[
\tilde{\mathbf{x}}
= \mathbf{x}_i
+ \lambda \bigl(\mathbf{x}_{i,\mathrm{nn}} - \mathbf{x}_i\bigr),
\quad
\lambda \sim \mathcal{U}(0,1).
\]
SMOTE became the de facto method for handling class imbalance, spawning many variations. Understanding SMOTE's interpolation principle is essential as it forms the foundation for regression adaptations like SMOTER and SMOGN that we compare against.

Along with SMOTE and its variations, advanced techniques such as generative adversarial networks (GANs) were extensively explored in the class imbalance literature. \citet{mariani2018bagan} developed BAGAN, a class-conditional GAN initialized with an autoencoder to produce diverse, high-quality minority images. \citet{tanaka2019dataaugmentation} demonstrated that GAN-generated tabular data could replace real samples in classifier training and improve minority recall. \citet{engelmann2020conditional} designed a conditional WGAN with gradient penalty, auxiliary classifier loss, Gumbel-softmax for categorical features, and cross-layer interactions for mixed-type tables. \citet{sharma2022smotified} embedded SMOTE-interpolated points into a GAN generator to diversify synthetic minority examples for addressing class imbalance. \citet{jiang2019gan} employed an encoder-decoder-encoder GAN on time-series data to detect anomalies via combined reconstruction and adversarial losses. \citet{lee2019ganids} generated rare-attack intrusion samples with a GAN and retrained a random forest to boost detection.

\subsection{Imbalanced regression}

Unlike class imbalance, the imbalanced regression problem received relatively little attention. \citet{torgo2007utility} and \citet{torgo2013} introduced the first data-level method (SMOTER) by extending SMOTE to regression with a relevance function $\phi(y)$ to identify rare targets, undersample common values, and interpolate rare points. \citet{branco2017} generalized SMOTER by adding Gaussian noise (SMOGN) to increase sample variability, and later proposed WERCS \citep{branco2019}, which probabilistically oversampled or undersampled based on $\phi(y)$ and simple random or noise generation. \citet{alahyari2025ldao} proposed LDAO, which clusters the joint feature-target space and applies kernel density estimation within each cluster to generate synthetic samples that preserve local distribution characteristics. \citet{camacho2022} introduced G-SMOTE, incorporating target-space geometry into interpolation, and \citet{camacho2024} replaced hard rarity thresholds with instance-based weighting in WSMOTER. \citet{alahyari2025smogan} developed SMOGAN, a two-stage framework that refines initial synthetic samples through adversarial training with distribution-aware GANs to improve sample quality. \citet{moniz2018} integrated SMOTE oversampling into boosting (SMOTEBoost), demonstrating the adaptability of data-level strategies for regression imbalance.

Alongside data-level strategies, several algorithmic methods were proposed for imbalanced regression. Cost-sensitive learning extended standard regression by weighting errors on rare targets \citep{zhou2010,elkan2001,domingos1999}. \citet{steininger2021} introduced DenseLoss, which emphasized low-density targets in the loss function. \citet{yang2021} applied label-distribution smoothing in deep neural networks. \citet{ren2022} proposed Balanced MSE to scale error contributions according to target rarity. 

There has also been work on developing new evaluation metrics specifically for imbalanced regression problems. \citet{ribeiro2020} developed SERA (Squared Error Relevance Area), a squared-error relevance-weighted metric that addresses the limitations of traditional metrics like RMSE in imbalanced settings. SERA integrates prediction errors across different relevance thresholds, giving more weight to errors on rare target values. This is particularly important because standard regression metrics often mask poor performance on minority samples by being dominated by the abundant majority samples, making specialized metrics like SERA essential for properly evaluating imbalanced regression methods.

\subsection{Mahalanobis distance and Gaussian mixture modeling in classification}
The Mahalanobis distance measures how far a data point lies from the center of a distribution, accounting for the correlation structure of the data. Unlike Euclidean distance, it takes into account the shape and orientation of the data cloud. This makes it particularly useful for outlier detection and understanding data geometry, which is why we employ it as the foundation of our minority detection approach. We provide formal definitions in Section 3.2.

Recent advances have applied Mahalanobis distance and Gaussian mixture modeling (GMM) in classification. \citet{nezhadshokouhi2020} proposed MAHAKIL, a Mahalanobis-based oversampling method for software defect prediction, demonstrating that Mahalanobis distance can increase intra-class diversity while preserving minority structure. \citet{yao2021emdo} introduced EMDO, a Mahalanobis distance-based method that fits multiple ellipsoids using GMM clustering and MOPSO to generate diverse minority samples in multiclass imbalanced settings. In the medical domain, \citet{balakrishnan2023mmote} developed MMOTE, a Mahalanobis metric-based oversampler that improves minority classification in Parkinson's disease detection by generating realistic gait samples. For fault diagnosis in electric motors, \citet{ribeiro2023gmmmaha} combined GMM with Mahalanobis distance to identify mechanical defects, showing high diagnostic accuracy across multiple failure modes. 

While these works demonstrate the power of Mahalanobis-GMM methods in classification, their application in imbalanced regression remains unexplored. Our work fills this gap by adapting Mahalanobis-GMM-based detection to the continuous target setting, using it to drive GAN-based synthetic data generation for minority samples.

\section{Method}
\label{sec:method}

\subsection{Overview}\label{sec:overview}
\begin{figure*}[!t]
  \centering
  \includegraphics[
    width=\linewidth,
    height=0.4\textheight
  ]{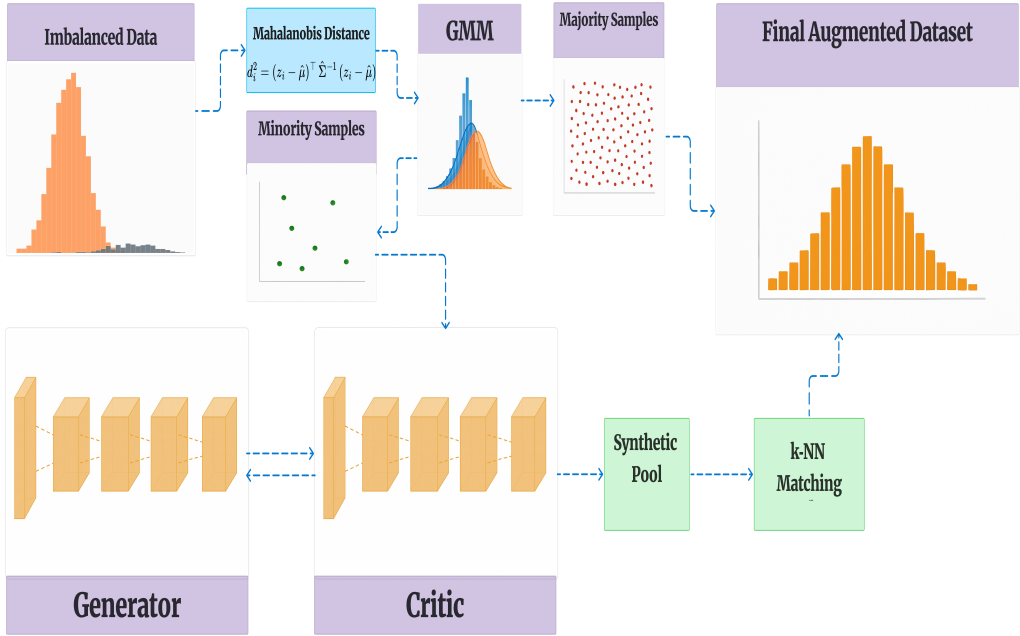}
  \caption{Architecture of the distance-based oversampling approach for imbalanced regression. The framework combines Mahalanobis distance metrics with Gaussian Mixture Models to generate synthetic samples for sparse target regions, using a Generator-Critic mechanism and k-NN matching to ensure synthetic sample quality and dataset balance.}
  \label{fig:mahalanobis}
\end{figure*}

Let $\mathbf x\in\mathbb R^{d}$ be the feature vector and $y\in\mathbb R$ the target; we stack them as $\mathbf z=(\mathbf x,y)\in\mathbb R^{p}$ with $p=d+1$. Our procedure comprises three sequential stages: (i) \emph{Mahalanobis-GMM detection}, which fits a univariate Gaussian mixture to the squared Mahalanobis distances of $\{\mathbf z_i\}$ and selects the high-distance component as minority samples; (ii) \emph{WGAN-GP generation}, which trains a Wasserstein GAN with gradient penalty on the detected minority set to produce a large synthetic pool; and (iii) \emph{deterministic matching}, which uses a robust Mahalanobis metric to match and filter the synthetic pool, yielding realistic augmentations for downstream regression. Figure~\ref{fig:mahalanobis} illustrates the pipeline of our approach.

\subsection{Stage 1: Mahalanobis–GMM minority detection}
\label{ssec:stage1}

Let each observation be a joint vector $\mathbf{z} = (\mathbf{x}, y)\in\mathbb{R}^p$, assumed to follow a multivariate Gaussian distribution $\mathbf{z}\sim\mathcal{N}(\boldsymbol{\mu},\boldsymbol{\Sigma})$, where $\boldsymbol{\mu}$ is the mean vector and $\boldsymbol{\Sigma}$ the $p\times p$ covariance matrix. The (squared) Mahalanobis distance from $\mathbf{z}$ to the mean is given by \citep{aggarwal2017introduction,ghorbani2019mahalanobis,ververidis2008}:
\begin{align}
d^2(\mathbf{z})
&= (\mathbf{z}-\boldsymbol{\mu})^\top \boldsymbol{\Sigma}^{-1} (\mathbf{z}-\boldsymbol{\mu})\,.
\label{eq:pop_maha}
\end{align}

In practice, given the training set $\mathcal{D}_{\mathrm{tr}}=\{\mathbf{z}_i\}_{i=1}^n\subset\mathbb{R}^p$, we compute empirical distances using the sample mean and covariance:
\[
\hat{\boldsymbol{\mu}}
= \frac{1}{n}\sum_{i=1}^n \mathbf{z}_i,\quad
\hat{\boldsymbol{\Sigma}}
= \frac{1}{n-1}\sum_{i=1}^n (\mathbf{z}_i-\hat{\boldsymbol{\mu}})(\mathbf{z}_i-\hat{\boldsymbol{\mu}})^\top,
\]
\begin{align}
d_i^2
&= (\mathbf{z}_i-\hat{\boldsymbol{\mu}})^\top \hat{\boldsymbol{\Sigma}}^{-1} (\mathbf{z}_i-\hat{\boldsymbol{\mu}}),\quad
i=1,\dots,n.
\label{eq:emp_maha}
\end{align}

Following \citet{Goodfellow-et-al-2016}, Gaussian mixture models are universal approximators of continuous densities.  A classical result is that if $\mathbf{z}\sim\mathcal{N}_p(\boldsymbol{\mu},\Sigma)$ then $d^2$ follows a chi-square distribution with $p$ degrees of freedom \citep{manly2005multivariate}. Since the chi-square distribution is heavy-tailed, we model $\{d_i^2\}$ with a two-component GMM, one component capturing the central (majority) bulk, the other capturing the tail (minority). Even if $\mathbf{z}$ is non-Gaussian, this Mahalanobis-GMM combination yields a robust, data-driven heuristic for detecting minority samples in the joint feature-target space.

We fit a two‐component Gaussian mixture to the squared distances $\{d_i^2\}$:
\begin{align}
p(x)
&= \underbrace{{\pi}\,\mathcal{N}\bigl(x\mid{\mu}_1,{\sigma}_1^2\bigr)}_{\text{minority}}
+\underbrace{(1-{\pi})\,\mathcal{N}\bigl(x\mid{\mu}_2,{\sigma}_2^2\bigr)}_{\text{majority}}
\end{align}
where  $\pi$ is the mixing weight. We estimate the two‐component GMM parameters via Maximum Likelihood (ML) using the EM algorithm \citep{mclachlan2008algorithm}. In practice, EM fitting of two Gaussians flexibly captures the central bulk $(\mu_1,\sigma_1^2)$ and any heavier tails $(\mu_2,\sigma_2^2)$. The resulting ML cutoff $T$ thus provides a principled, data‐driven separator between majority and minority distances by comparing each $d_i^2$ to $T$:
\[
\begin{cases}
d_i^2 < T:&\text{assign to the majority component},\\
d_i^2 \ge T:&\text{assign to the minority component}.
\end{cases}
\] 
The cutoff $T$ is defined as the unique solution of the weighted‐density equality
\begin{align}
\hat{\pi}\,\mathcal{N}\bigl(T\mid\hat{\mu}_1,\hat{\sigma}_1^2\bigr)
&= (1-\hat{\pi})\,\mathcal{N}\bigl(T\mid\hat{\mu}_2,\hat{\sigma}_2^2\bigr),
\label{eq:threshold}
\end{align}
so that $x=T$ has equal likelihood under both components. 
After taking logarithms and rearranging, this intersection condition yields the quadratic equation
\begin{align}
(\hat{\sigma}_1^2-\hat{\sigma}_2^2)\,T^2
+(-2\hat{\sigma}_1^2\hat{\mu}_2+2\hat{\sigma}_2^2\hat{\mu}_1)\,T\notag\\
+\hat{\sigma}_1^2\hat{\mu}_2^2-\hat{\sigma}_2^2\hat{\mu}_1^2
-2\hat{\sigma}_1^2\hat{\sigma}_2^2
  \ln\!\frac{(1-\hat{\pi})\,\hat{\sigma}_1}{\hat{\pi}\,\hat{\sigma}_2}
=0,
\label{eq:quad}
\end{align}
Solving this quadratic equation provides a fully automatic, statistically principled seed for the subsequent GAN refinement stage.

\subsection{Stage 2: Minority sample generation with WGAN-GP}
\label{ssec:stage2}

We train a Wasserstein GAN with gradient penalty (WGAN-GP) \citep{arjovsky2017wasserstein,gulrajani2017improved} on the minority set \(\mathcal{D}_{\min}\subset\mathbb{R}^{p}\), defined by the value T from the previous section. We choose WGAN-GP because it provides stable training dynamics and generates diverse, high-quality samples without mode collapse, critical for preserving the complex structure of minority regions in tabular data. Let the generator \(G:\mathbb{R}^{q}\!\rightarrow\!\mathbb{R}^{p}\) and critic \(D:\mathbb{R}^{p}\!\rightarrow\!\mathbb{R}\) be multilayer perceptrons with ReLU activations.

\noindent{Critic loss:}
\begin{align}
\mathcal{L}_{D}
 &=\;
    \mathbb{E}_{\,\boldsymbol{\epsilon}\sim\mathcal{N}(\mathbf{0},I_{q})}
      \bigl[D\bigl(G(\boldsymbol{\epsilon})\bigr)\bigr]
  - \mathbb{E}_{\,\mathbf{z}\in\mathcal{D}_{\min}}
      \bigl[D(\mathbf{z})\bigr]                                       \notag\\
 &\quad
  +\lambda_{\mathrm{gp}}
    \mathbb{E}_{\,\widehat{\mathbf{u}}}
      \Bigl(\lVert\nabla_{\widehat{\mathbf{u}}}D(\widehat{\mathbf{u}})\rVert_{2}-1\Bigr)^{2},
\label{eq:critic}
\end{align}
where \(\widehat{\mathbf{u}}=\alpha\mathbf{z}+(1-\alpha)G(\boldsymbol{\epsilon})\)
with \(\alpha\sim\mathcal{U}(0,1)\).

\noindent{Generator loss:}
\begin{align}
\mathcal{L}_{G}
  = -\,\mathbb{E}_{\,\boldsymbol{\epsilon}\sim\mathcal{N}(\mathbf{0},I_{q})}
      \bigl[D\bigl(G(\boldsymbol{\epsilon})\bigr)\bigr].
\label{eq:gen}
\end{align}

After convergence, we draw $N_{\mathrm{pool}}$ synthetic samples from the generator:
\[
  \mathcal{D}_{\mathrm{syn}}^{\mathrm{pool}}
  = \bigl\{G(\boldsymbol{\epsilon}_{j})\bigr\}_{j=1}^{N_{\mathrm{pool}}},
  \quad
  \boldsymbol{\epsilon}_{j}\sim\mathcal{N}(\mathbf{0},I_{q}).
\]

\subsection{Stage 3: Deterministic Nearest-Neighbour matching}
\label{ssec:stage3}

We employ a k-nearest neighbour (k-NN) approach to select high-quality synthetic samples from the generated pool, ensuring they closely match the distribution of real minority points. Let \(\mathbf{V} = \hat{\boldsymbol{\Sigma}}^{-1}\) be the precision matrix from Stage 1 (Eq.~\eqref{eq:pop_maha}).  
While earlier distances were measured from the mean, we now compute pairwise Mahalanobis distances between real and synthetic points.  
For each real minority point \(\mathbf{z}_{r} \in \mathcal{D}_{\min}\) and each synthetic candidate \(\tilde{\mathbf{z}} \in \mathcal{D}_{\mathrm{syn}}^{\mathrm{pool}}\), we compute
\begin{equation}
\delta\!\bigl(\mathbf{z}_{r},\tilde{\mathbf{z}}\bigr)
  = \sqrt{%
     \bigl(\tilde{\mathbf{z}} - \mathbf{z}_{r}\bigr)^{\!\top}
     \mathbf{V}\,
     \bigl(\tilde{\mathbf{z}} - \mathbf{z}_{r}\bigr)}.
\label{eq:match_dist}
\end{equation}

For each \(\mathbf{z}_{r}\) sort the pool by
\(\delta(\mathbf{z}_{r},\cdot)\) and keep the
\(k\) nearest candidates:  
\[
  \mathcal{N}\!\left(\mathbf{z}_{r}\right)
    = \bigl\{\tilde{\mathbf{z}}_{(1)},\dots,
              \tilde{\mathbf{z}}_{(k)}\bigr\},
\]
where \(\delta(\mathbf{z}_{r},\tilde{\mathbf{z}}_{(1)})\le
       \cdots\le
       \delta(\mathbf{z}_{r},\tilde{\mathbf{z}}_{(k)})\).



Next, select the \(n_{\mathrm{pick}} \le k\) closest candidates from \(\mathcal{N}(\mathbf{z}_{r})\), where \(n_{\mathrm{pick}}\) is a hyperparameter controlling sample diversity:
\[
  \mathcal{M}\!\left(\mathbf{z}_{r}\right)
    = \bigl\{\tilde{\mathbf{z}}_{(1)},\dots,
              \tilde{\mathbf{z}}_{(n_{\mathrm{pick}})}\bigr\}.
\] 
We then define the final refined synthetic set as the union over all real minority points:
\begin{equation}
  \mathcal{D}_{\mathrm{syn}}^{\mathrm{ref}}
  = \bigcup_{\mathbf{z}_{r}\in\mathcal{D}_{\min}}
      \mathcal{M}\!\left(\mathbf{z}_{r}\right).
\label{eq:d_syn_ref}
\end{equation}

\subsection{Final augmented dataset}
\label{ssec:final}

The data presented to the downstream regressor are
\[
  \mathcal{D}_{\mathrm{train}}^{\mathrm{aug}}
    = \mathcal{D}_{\mathrm{tr}}
      \;\cup\;
      \mathcal{D}_{\mathrm{syn}}^{\mathrm{ref}}.
\]
This union enriches sparse regions while keeping all original samples
intact.

\subsection{Hyperparameters and architecture}
\label{ssec:params}

Our default settings, which can be tuned in practice, are as follows:
the generator uses a
\(q \!\to\! 128 \!\to\! 256 \!\to\! 128 \!\to\! p\) architecture with ReLU
activations; the critic uses a
\(p \!\to\! 256 \!\to\! 256 \!\to\! 128 \!\to\! 1\) architecture with
ReLU activations; the latent-noise dimension is \(q = 64\); the
gradient-penalty coefficient is \(\lambda_{\mathrm{gp}} = 10\); the
synthetic-pool size is \(N_{\mathrm{pool}} = 10{,}000\); matching first
collects the \(k = 3\) nearest candidates and then keeps the
\(n_{\mathrm{pick}} = 1\) closest unused one for each real minority
point.

\section{Evaluation methodology}
We benchmark our framework against current state-of-the-art imbalanced regression oversamplers SMOGN \citep{branco2017} and G-SMOTE \citep{camacho2022}, as well as random oversampling (RO) \citep{he2009} and a baseline model without any oversampling. We evaluated our method using 32 datasets from the Keel repository \citep{alcala2011} and \citep{branco2019}. Table~\ref{tab:dataset-characteristics} shows the number of instances and features for each dataset.

\begin{table}[t]
  \centering
  \scriptsize
  \setlength{\tabcolsep}{3pt}
  \caption{Dataset characteristics showing number of instances and features.}
  \label{tab:dataset-characteristics}
  \resizebox{\columnwidth}{!}{%
  \begin{tabular}{lrr | lrr}
    \toprule
    dataset           & instances & features & dataset           & instances & features \\
    \midrule
    a1                & 198       & 11       & debutanizer       & 2,394     & 7        \\
    a2                & 198       & 11       & ele-2             & 1,056     & 4        \\
    a3                & 198       & 11       & forestfires       & 517       & 12       \\
    a7                & 198       & 11       & fuel              & 1,764     & 37       \\
    abalone           & 4,177     & 8        & heat              & 7,400     & 11       \\
    acceleration      & 1,732     & 14       & house             & 22,784    & 16       \\
    airfoild          & 1,503     & 5        & kdd               & 316       & 18       \\
    analcat           & 450       & 11       & laser             & 993       & 4        \\
    available\_power  & 1,802     & 15       & maximal\_torque   & 1,802     & 32       \\
    baseball          & 337       & 16       & meta              & 528       & 65       \\
    boston            & 506       & 13       & mortgage          & 1,049     & 15       \\
    california        & 20,640    & 8        & sensory           & 576       & 11       \\
    compactiv         & 8,192     & 21       & treasury          & 1,049     & 15       \\
    concrete\_strength& 1,030     & 8        & triazines         & 186       & 60       \\
    cpu               & 8,192     & 12       & wine\_quality     & 1,143     & 12       \\
    lungcancer        & 442       & 24       & machineCPU        & 209       & 6        \\
    \bottomrule
  \end{tabular}%
  }
\end{table}

\subsection{Metrics}
We compare methods using RMSE, SERA and F$_{\phi_1}$:
\[
\mathrm{RMSE}
=\sqrt{\frac{1}{n}\sum_{i=1}^{n}(y_i-\hat y_i)^2}\,.
\]
SERA \citep{ribeiro2020} uses a relevance function $\phi:\mathcal Y\to[0,1]$, 
$D^t=\{(x_i,y_i)\mid\phi(y_i)\ge t\}$, 
$SER_t=\sum_{(x_i,y_i)\in D^t}(\hat y_i-y_i)^2$, and 
\[
\mathrm{SERA}
=\int_{0}^{1}SER_t\,dt
\;
\]
Regression precision, recall and F$_{\phi_1}$ employ threshold $t_R$ and utility $U(\hat y_i,y_i)\in[-1,1]$, where utility $U(\hat y_i, y_i)$ quantifies the usefulness of predicting $\hat y_i$ when the true value is $y_i$, based on the relevance of $y_i$ and the error between $\hat y_i$ and $y_i$ \citep{torgo2009, ribeiro2011a, branco2019}:
\begin{align*}
  \mathrm{prec}_\phi &= \frac{\sum_{\phi(\hat y_i)>t_R}\bigl(1 + U(\hat y_i,y_i)\bigr)}
                          {\sum_{\phi(\hat y_i)>t_R}\bigl(1 + \phi(\hat y_i)\bigr)},\\
  \mathrm{rec}_\phi  &= \frac{\sum_{\phi(y_i)>t_R}\bigl(1 + U(\hat y_i,y_i)\bigr)}
                          {\sum_{\phi(y_i)>t_R}\bigl(1 + \phi(y_i)\bigr)},\\
  F_{\phi_1}        &= \frac{2\,\mathrm{prec}_\phi\,\mathrm{rec}_\phi}
                          {\mathrm{prec}_\phi + \mathrm{rec}_\phi}.
\end{align*}

\subsection{Experimental framework}
We performed 25 random 80/20 train/test splits per dataset, using SERA and RO from ImbalancedLearningRegression \citep{wu2022} and SMOGN per \citep{kunz2020}. We trained TabNet \citep{arik2021tabnet}, holding out 20\% of training for validation, with Adam (learning rate 1e-4) for up to 1000 epochs, selecting the checkpoint with lowest validation RMSE. Our TabNet comprised six transformer blocks (one initial plus five decision), each containing four 64-unit fully connected layers, five 16-unit attention layers, and one output neuron. We also trained a scikit-learn Random Forest \citep{breiman2001random} on the full training set and evaluated it on the test split.

\section{Qualitative evaluation}

In this section, using the Boston dataset as a representative example, we qualitatively evaluate how our framework captures minority samples in the joint feature-target distribution and assess the quality of the matched nearest neighbours drawn from the synthetic pool generated by our WGAN. The split value determined is $T=20.13$, which represents the Mahalanobis distance threshold.

\begin{figure}[!t]
  \centering
  \begin{minipage}[t]{0.44\linewidth}
    \centering
    \includegraphics[width=\linewidth]{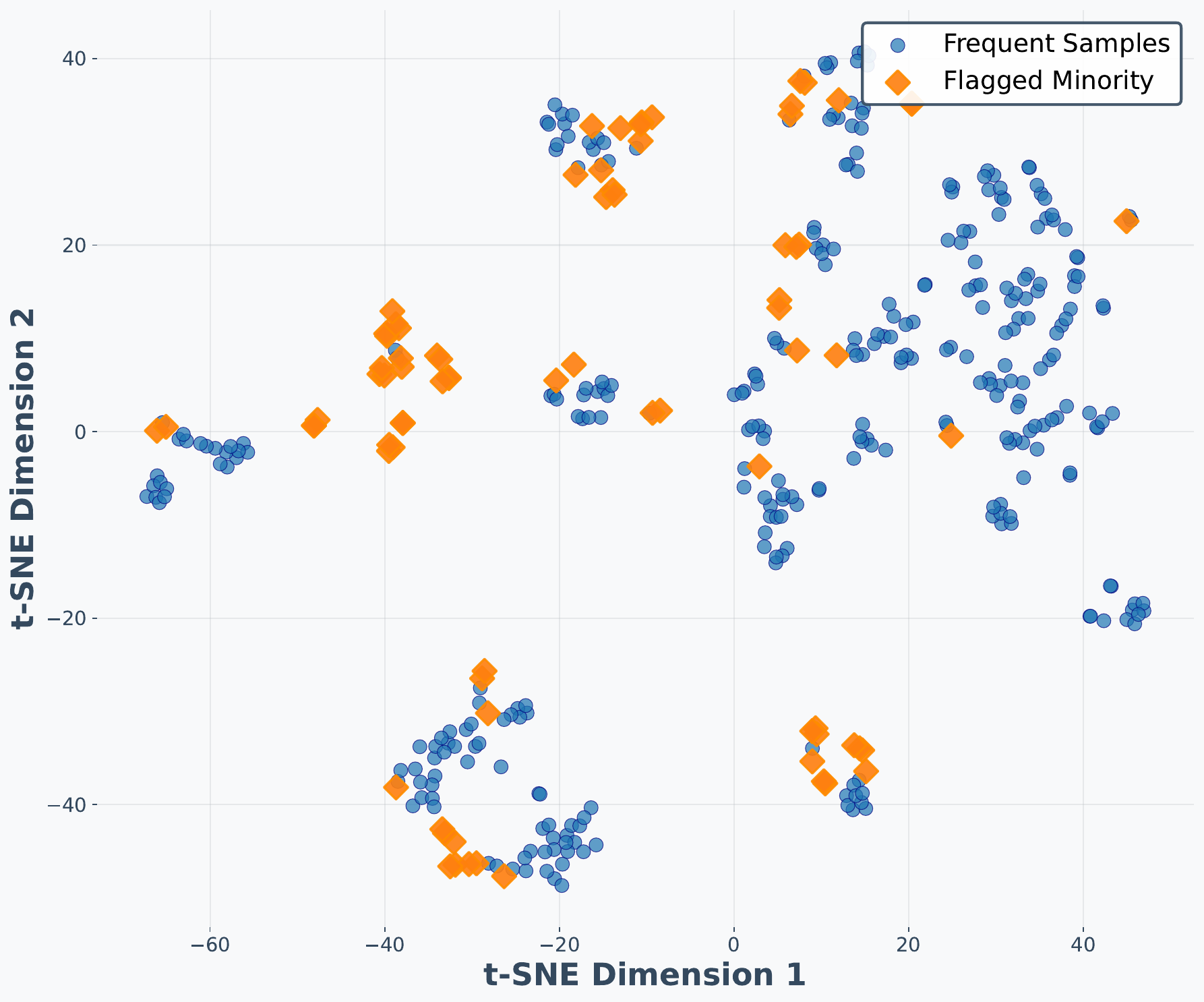}
    \par\smallskip
    {\scriptsize\textbf{(a)} Perplexity = 10}
  \end{minipage}
  \begin{minipage}[t]{0.44\linewidth}
    \centering
    \includegraphics[width=\linewidth]{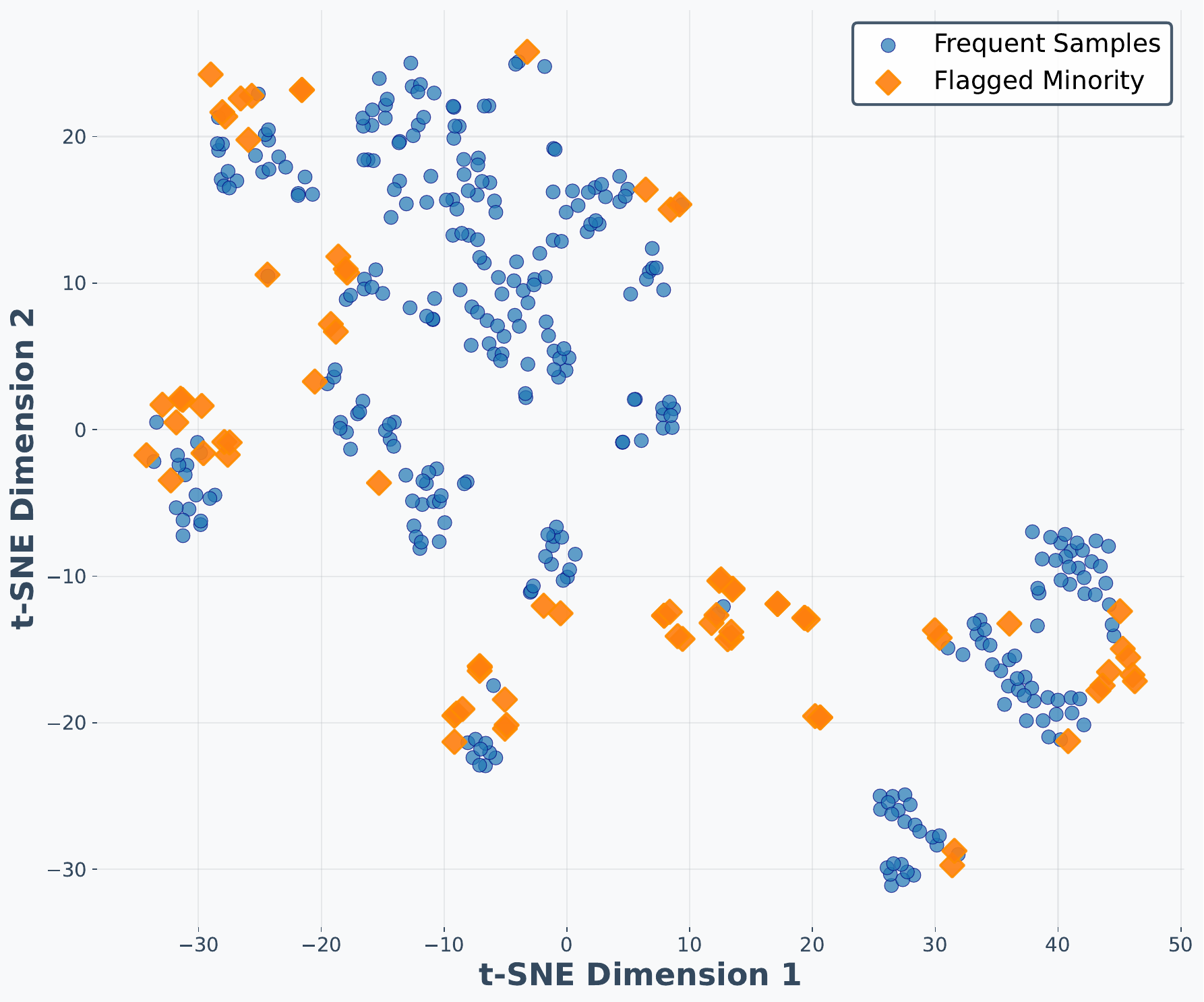}
    \par\smallskip
    {\scriptsize\textbf{(b)} Perplexity = 20}
  \end{minipage}
  \begin{minipage}[t]{0.44\linewidth}
    \centering
    \includegraphics[width=\linewidth]{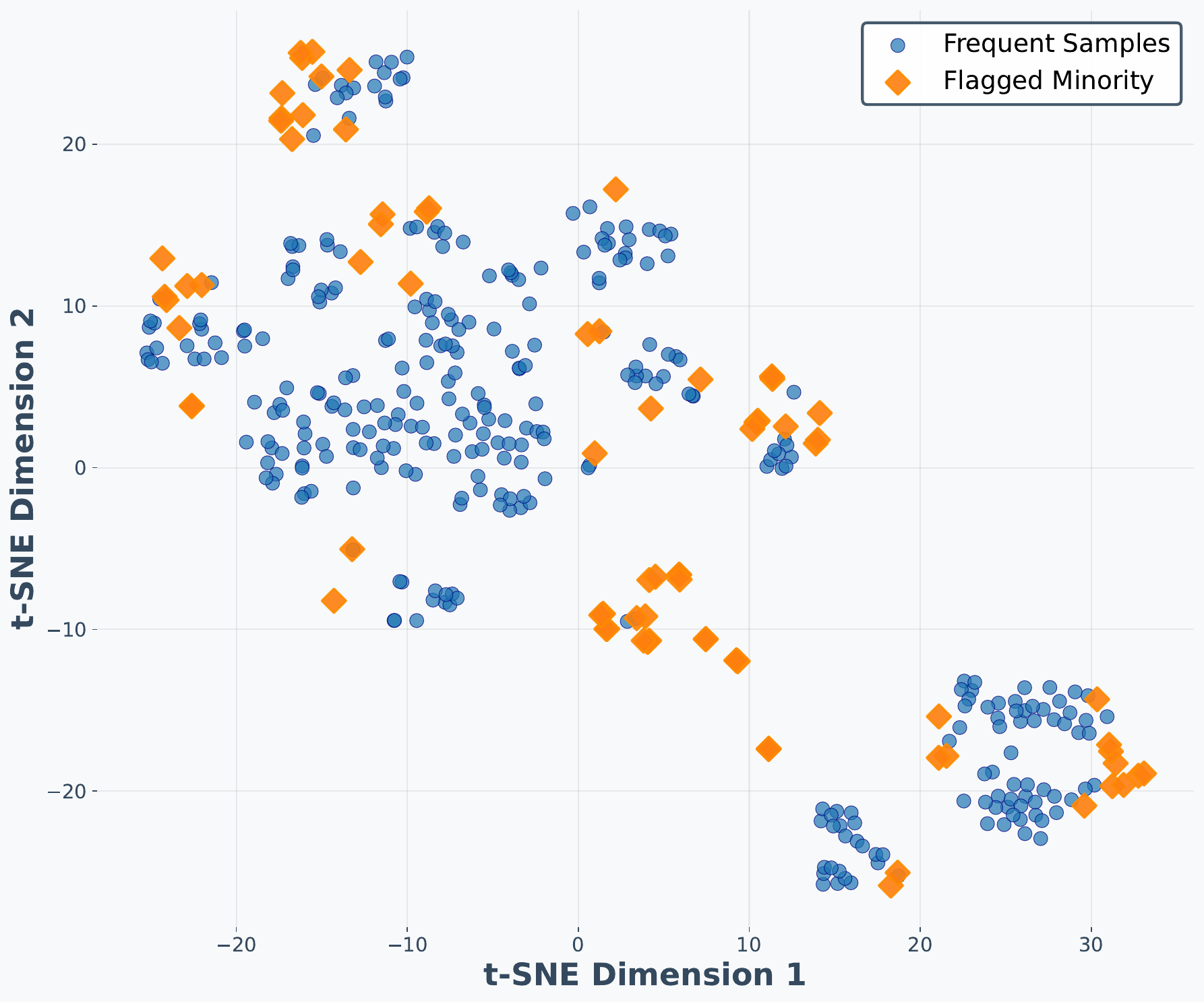}
    \par\smallskip
    {\scriptsize\textbf{(c)} Perplexity = 30}
  \end{minipage}
  \begin{minipage}[t]{0.44\linewidth}
    \centering
    \includegraphics[width=\linewidth]{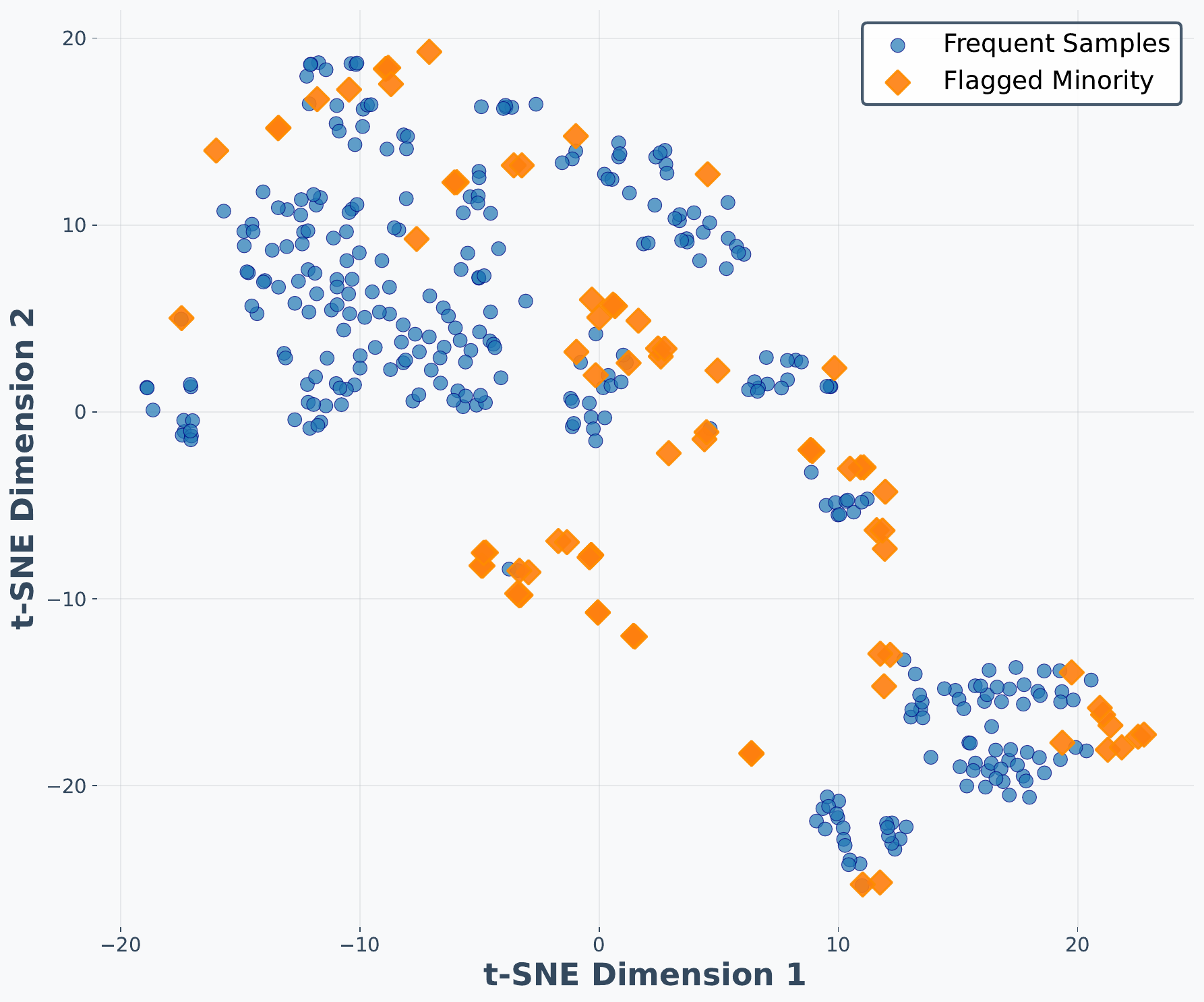}
    \par\smallskip
    {\scriptsize\textbf{(d)} Perplexity = 40}
  \end{minipage}
  \caption{t-SNE visualization with different perplexity values.}
  \label{fig:tsne_perplexity_comparison}
\end{figure}

\subsection{t-distributed stochastic neighbour embedding}

t-SNE is a nonlinear dimensionality reduction technique that converts high-dimensional pairwise similarities into probabilities and finds a low-dimensional embedding by minimizing the KL divergence \citep{vandermaaten2008visualizing}. Figure~\ref{fig:tsne_perplexity_comparison} shows Boston dataset embeddings with majority samples as blue circles and minority samples specified by our method as orange diamonds at perplexities 10, 20, 30 and 40. At perplexity 10, local neighbourhoods dominate, fragmenting minority points; at 20, local and global structure balance, forming coherent minority clusters; at 30, small fluctuations are smoothed while preserving cluster separation; at 40, global organisation is most pronounced but fine-grained patterns merge. These results demonstrate that our method reliably captures the true minority structure across both local and global scales.

\begin{figure}[!t]
  \centering
  \includegraphics[width=0.8\linewidth]{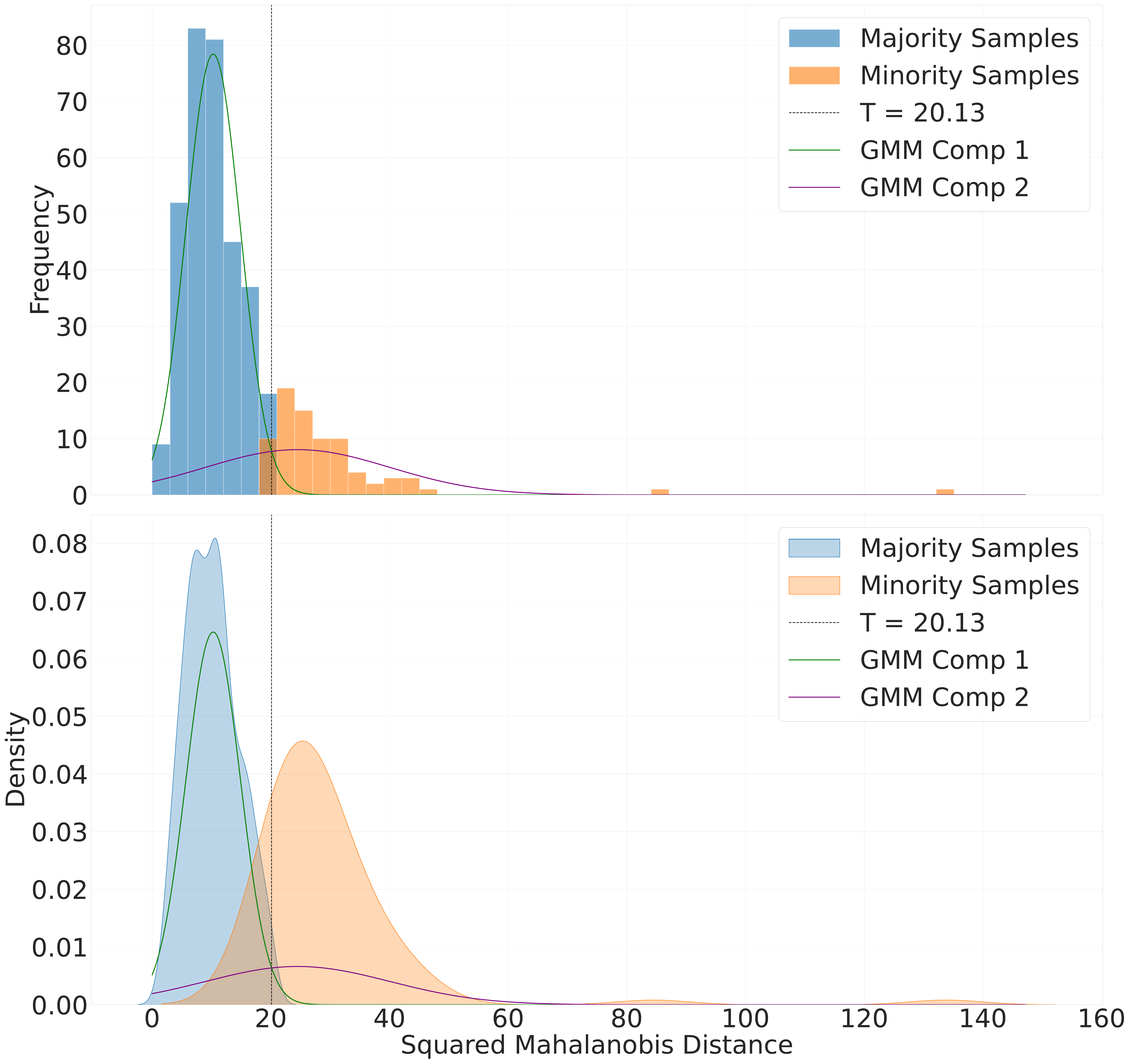}
  \caption{Mahalanobis-GMM analysis on the Boston dataset. Top: frequency histogram of squared Mahalanobis distances for majority (blue) and flagged minority (orange) samples, overlaid with the two GMM component curves (green and purple) and the intersection threshold \(T\). Bottom: corresponding kernel density estimates with the same color scheme and threshold.}
  \label{fig:mahalanobis_gmm}
\end{figure}

\begin{figure}[!t]
  \centering
  \includegraphics[width=0.8\linewidth]{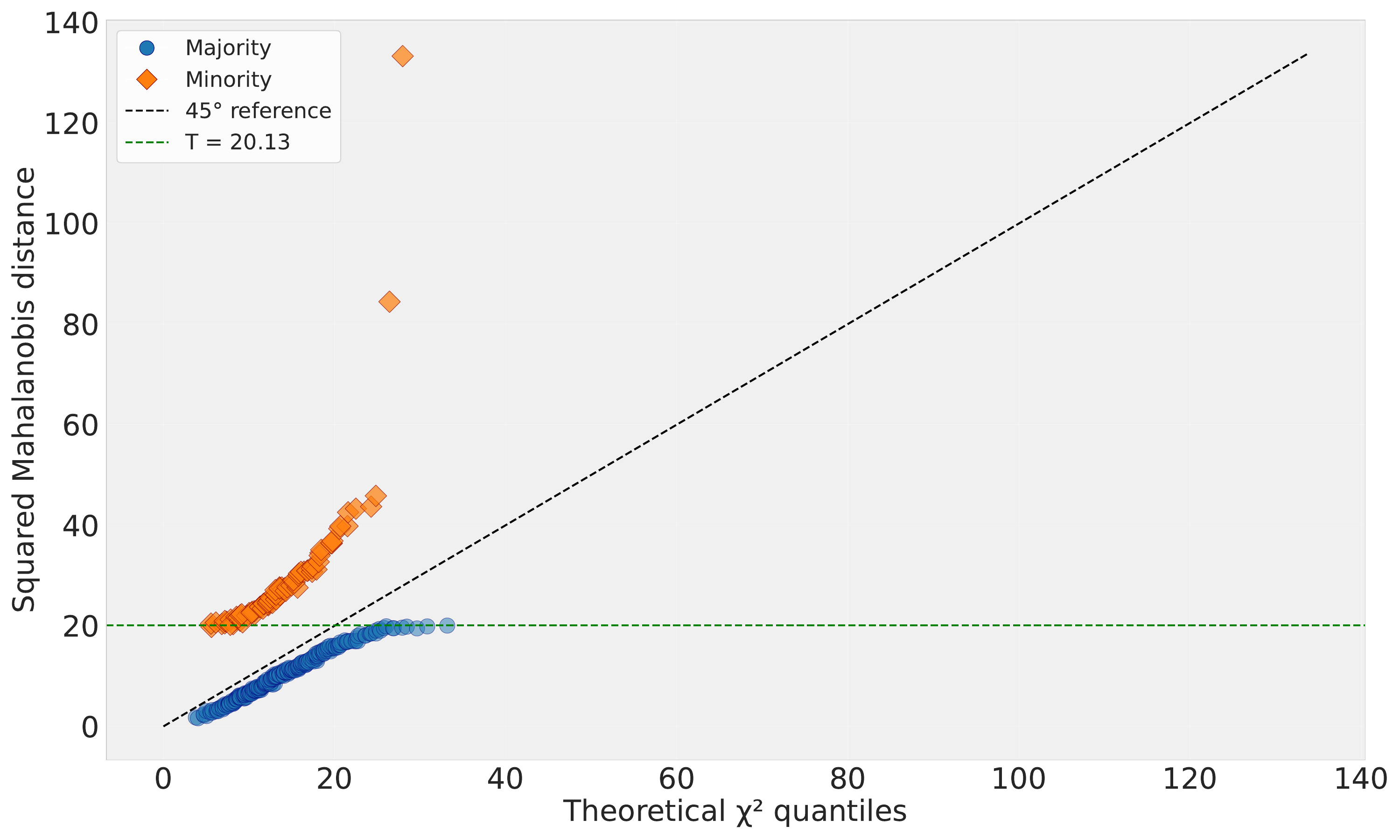}
  \caption{Q-Q plot of squared Mahalanobis distances vs.\ theoretical \(\chi^2_p\) quantiles. Blue circles: majority samples; orange diamonds: detected minority samples; black dashed: 45° reference line; green dashed: threshold \(T\).}
  \label{fig:mahalanobis_qq}
\end{figure}

\begin{figure}[!t]
  \centering
  \begin{minipage}[t]{0.44\linewidth}
    \centering
    \includegraphics[width=\linewidth]{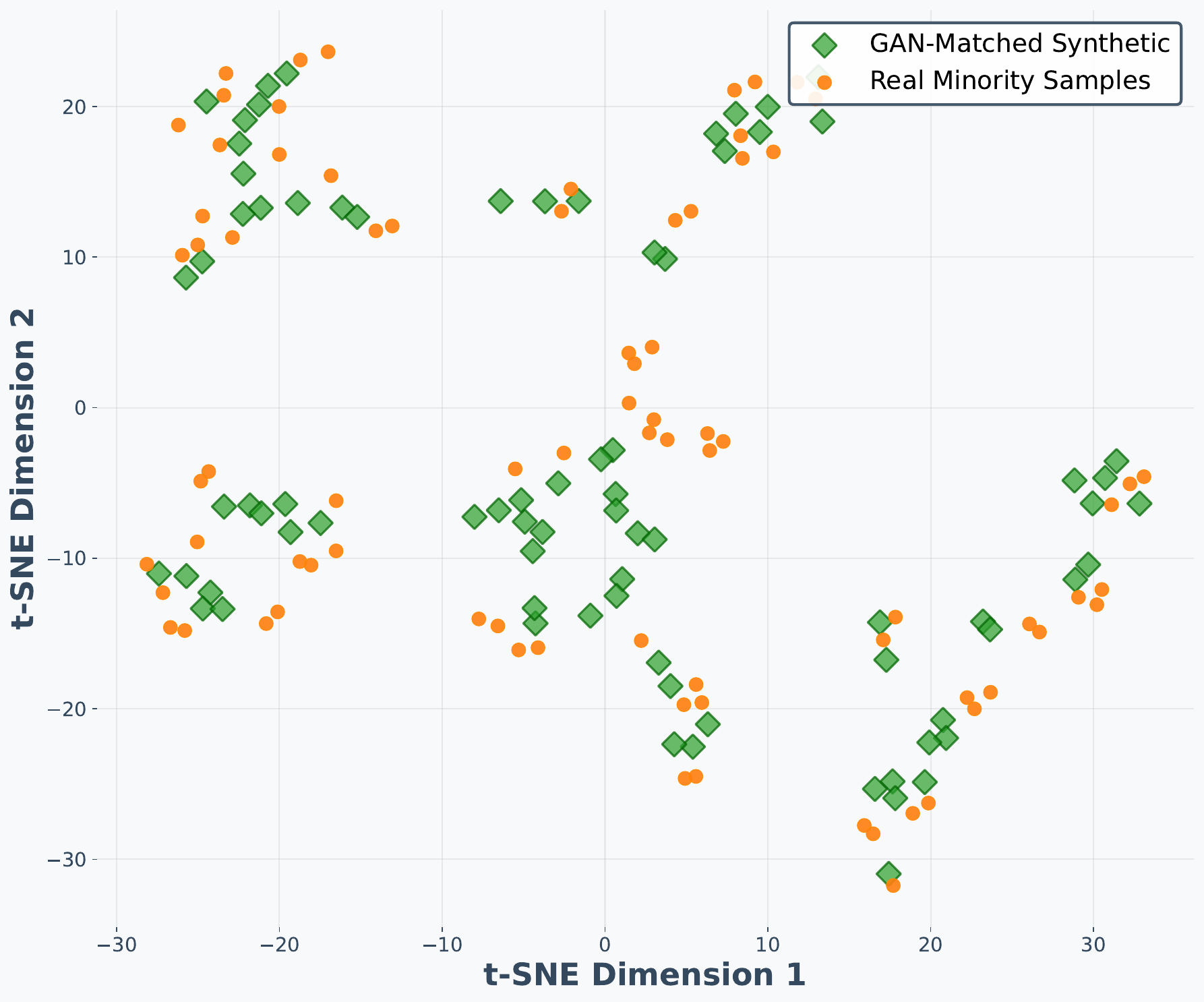}
    \par\smallskip
    {\scriptsize\textbf{(a)} Perplexity = 10}
    \label{fig:tsne_min_synth_p10}
  \end{minipage}
  \begin{minipage}[t]{0.44\linewidth}
    \centering
    \includegraphics[width=\linewidth]{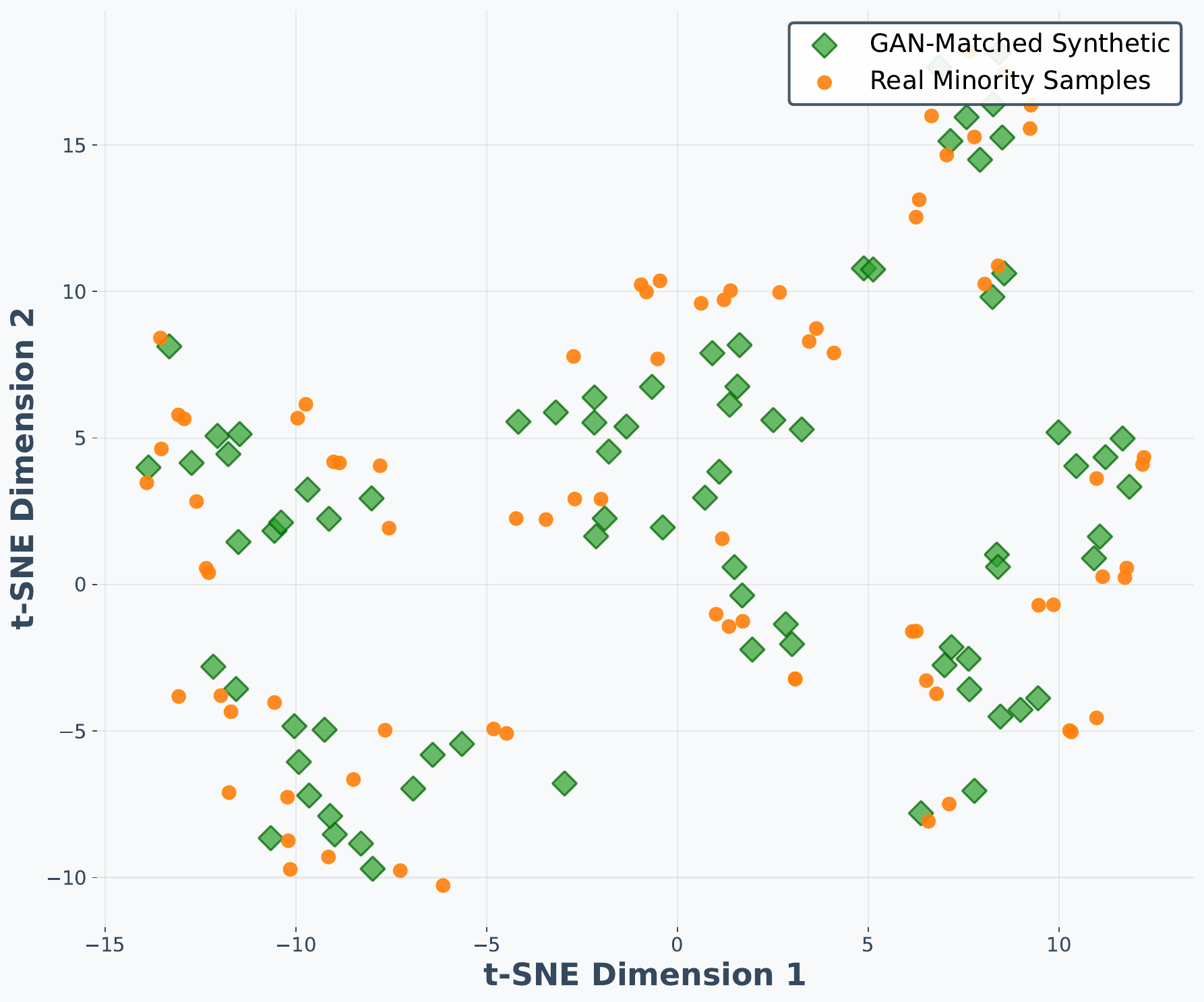}
    \par\smallskip
    {\scriptsize\textbf{(b)} Perplexity = 20}
    \label{fig:tsne_min_synth_p20}
  \end{minipage}
  \begin{minipage}[t]{0.44\linewidth}
    \centering
    \includegraphics[width=\linewidth]{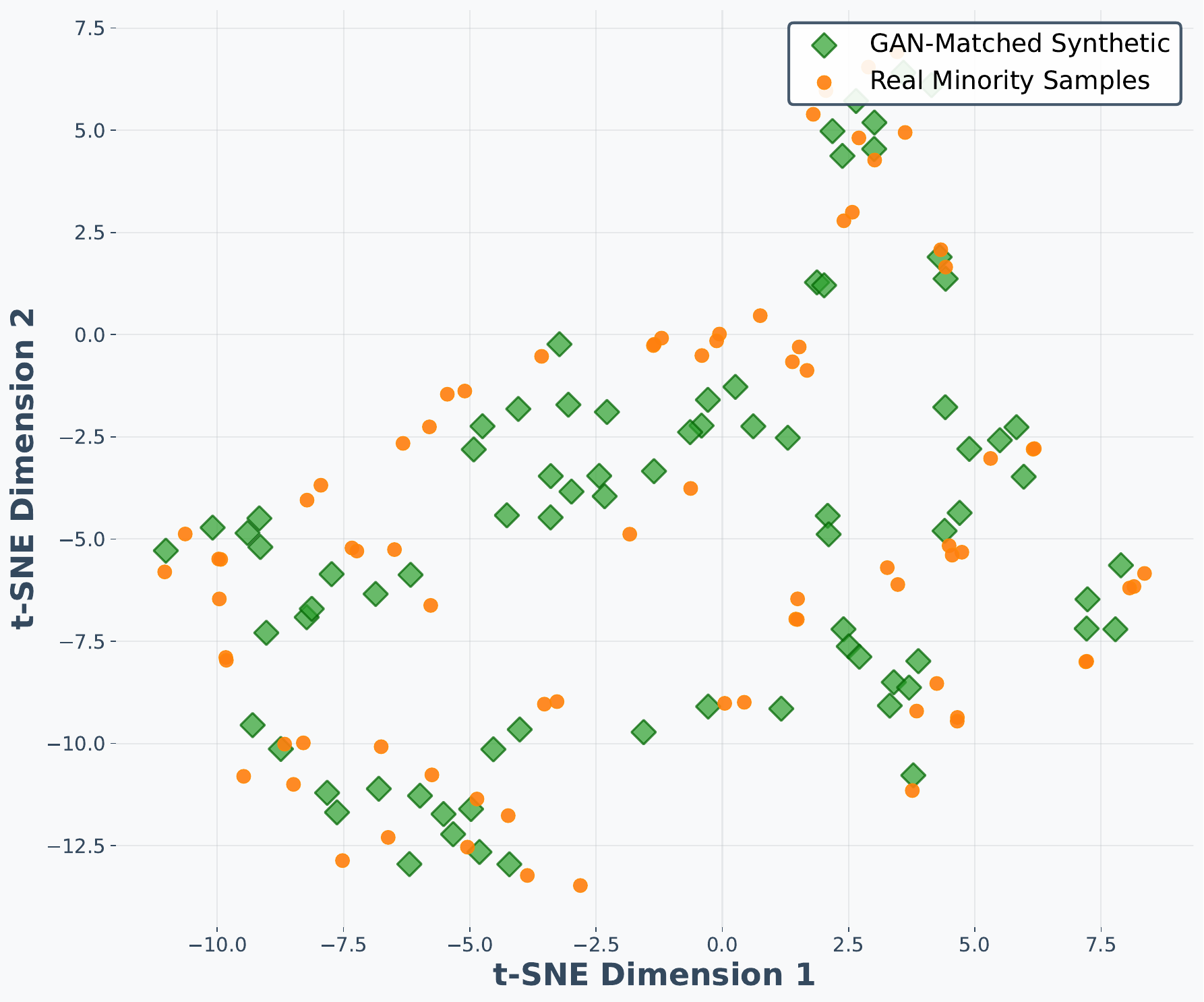}
    \par\smallskip
    {\scriptsize\textbf{(c)} Perplexity = 30}
    \label{fig:tsne_min_synth_p30}
  \end{minipage}
  \begin{minipage}[t]{0.44\linewidth}
    \centering
    \includegraphics[width=\linewidth]{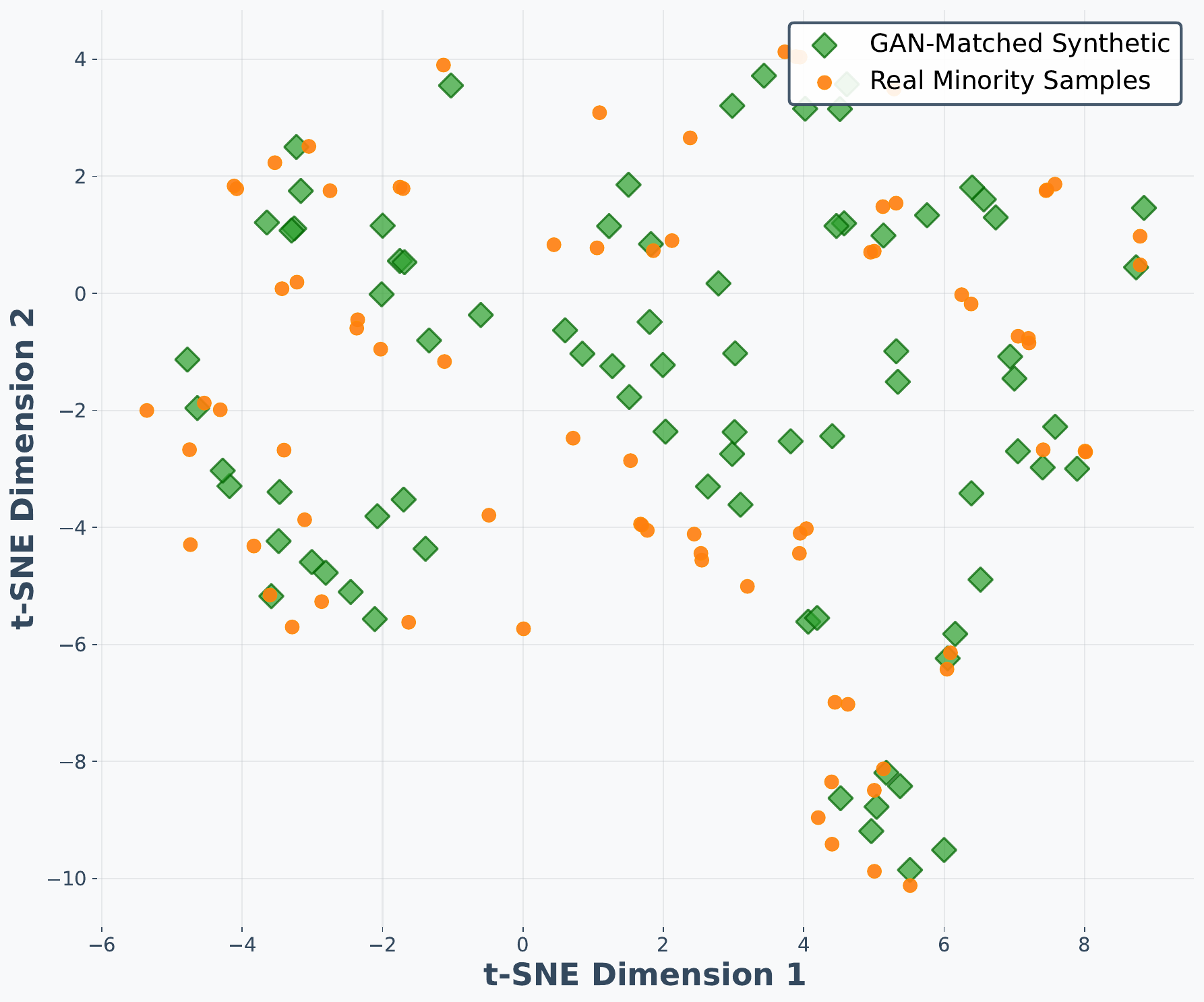}
    \par\smallskip
    {\scriptsize\textbf{(d)} Perplexity = 40}
    \label{fig:tsne_min_synth_p40}
  \end{minipage}
  \caption{t-SNE embeddings of real minority versus GAN-matched synthetic samples for perplexities of 10, 20, 30, and 40. Real minority samples are orange circles; GAN-matched synthetic samples are green diamonds. Close overlap in each panel confirms that matched synthetic points reproduce the true minority structure across scales.}
  \label{fig:tsne_minority_synth}
\end{figure}

\subsection{Assessment of GMM-based thresholding}
\label{ssec:maha_assessment}

The top panel of Figure~\ref{fig:mahalanobis_gmm} shows a clear gap in the histogram of squared Mahalanobis distances at the threshold \(T\). The majority samples clusters tightly to the left of \(T\), while the flagged minority samples form a distinct right‐hand tail with little overlap. This indicates that the intersection of the two fitted Gaussian components accurately pinpoints the boundary between common and rare observations.

The GMMs defined by Eq.~\eqref{eq:threshold} yield two distributions, which we plot in the lower panel of Figure~\ref{fig:mahalanobis_gmm}. In this panel, the majority density drops sharply to near zero at \(T\), and the minority density increases immediately thereafter. The low‐density valley between the two GMM curves aligns perfectly with our cutoff, confirming that the method avoids misclassifying central points as outliers and captures nearly all true extremes. By deriving the threshold directly from the data distribution, without arbitrary thresholds, this procedure adapts to the actual shape of the distance distribution. The result is a robust, automatic selection of genuinely rare samples, providing a reliable input set for the subsequent adversarial refinement stage.

\subsection{Chi-square Q-Q analysis of Mahalanobis distances}
\label{ssec:maha_qq}

Figure~\ref{fig:mahalanobis_qq} presents a Q-Q plot comparing the empirical squared Mahalanobis distances to the theoretical \(\chi^2_p\) distribution (with \(p\) degrees of freedom). Majority samples (blue circles) lie close to the 45° reference line up to the threshold \(T\), confirming that their distances follow the expected \(\chi^2_p\) law. In contrast, the flagged minority samples (orange diamonds) systematically exceed \(T\) in the upper tail, underscoring their role as genuine outliers. The horizontal green dashed line marks the GMM-derived cutoff \(T\), and deviations of the majority points from the reference line would signal possible misspecification of the covariance estimate.

\begin{figure*}[!t]
 \centering
 \begin{minipage}[t]{0.45\linewidth}
   \centering
   \includegraphics[width=\linewidth]{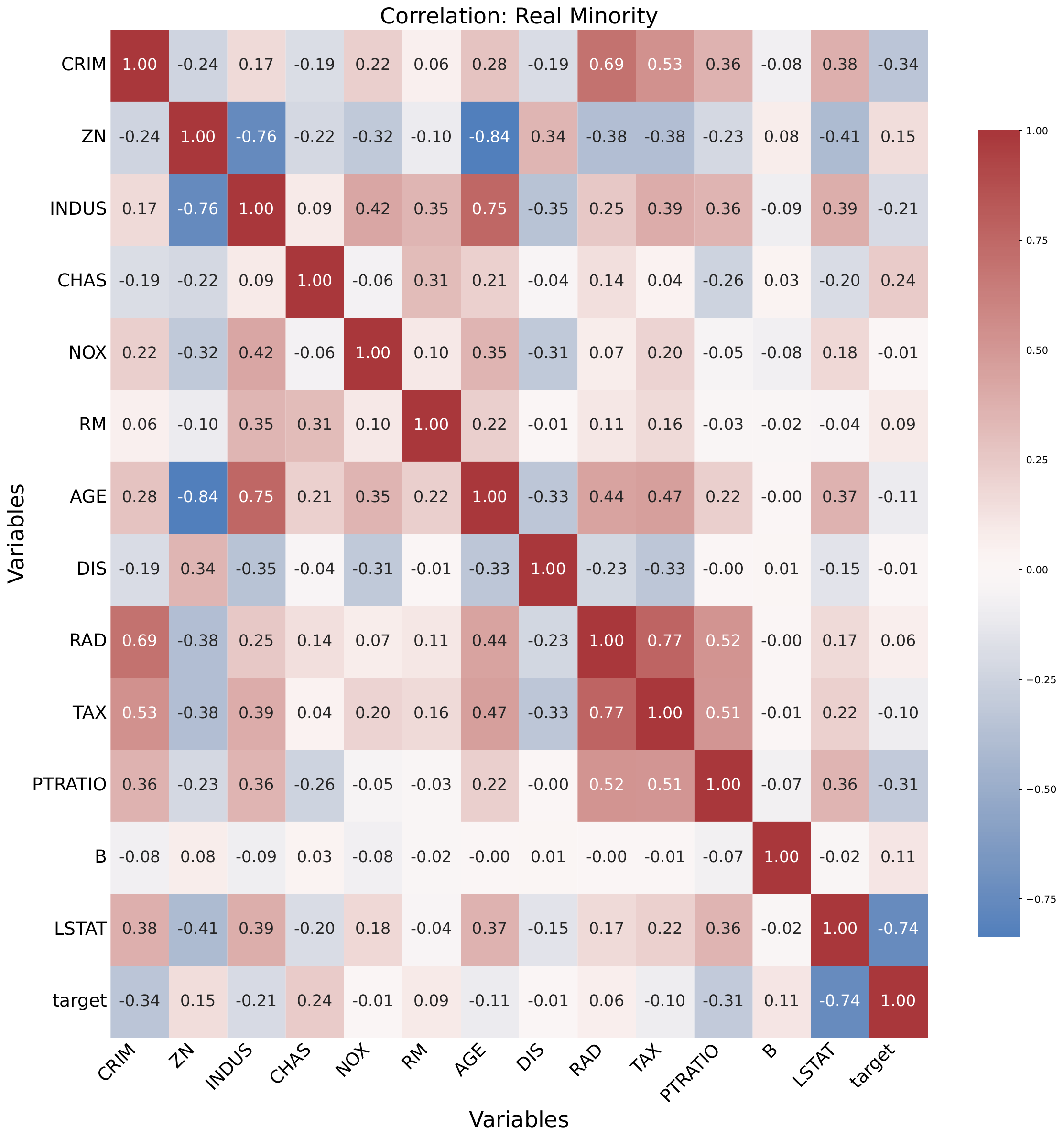}
   \par\smallskip
   {\scriptsize\textbf{(a)} Real minority correlations}
   \label{fig:corr_real}
 \end{minipage}\hfill
 \begin{minipage}[t]{0.45\linewidth}
   \centering
   \includegraphics[width=\linewidth]{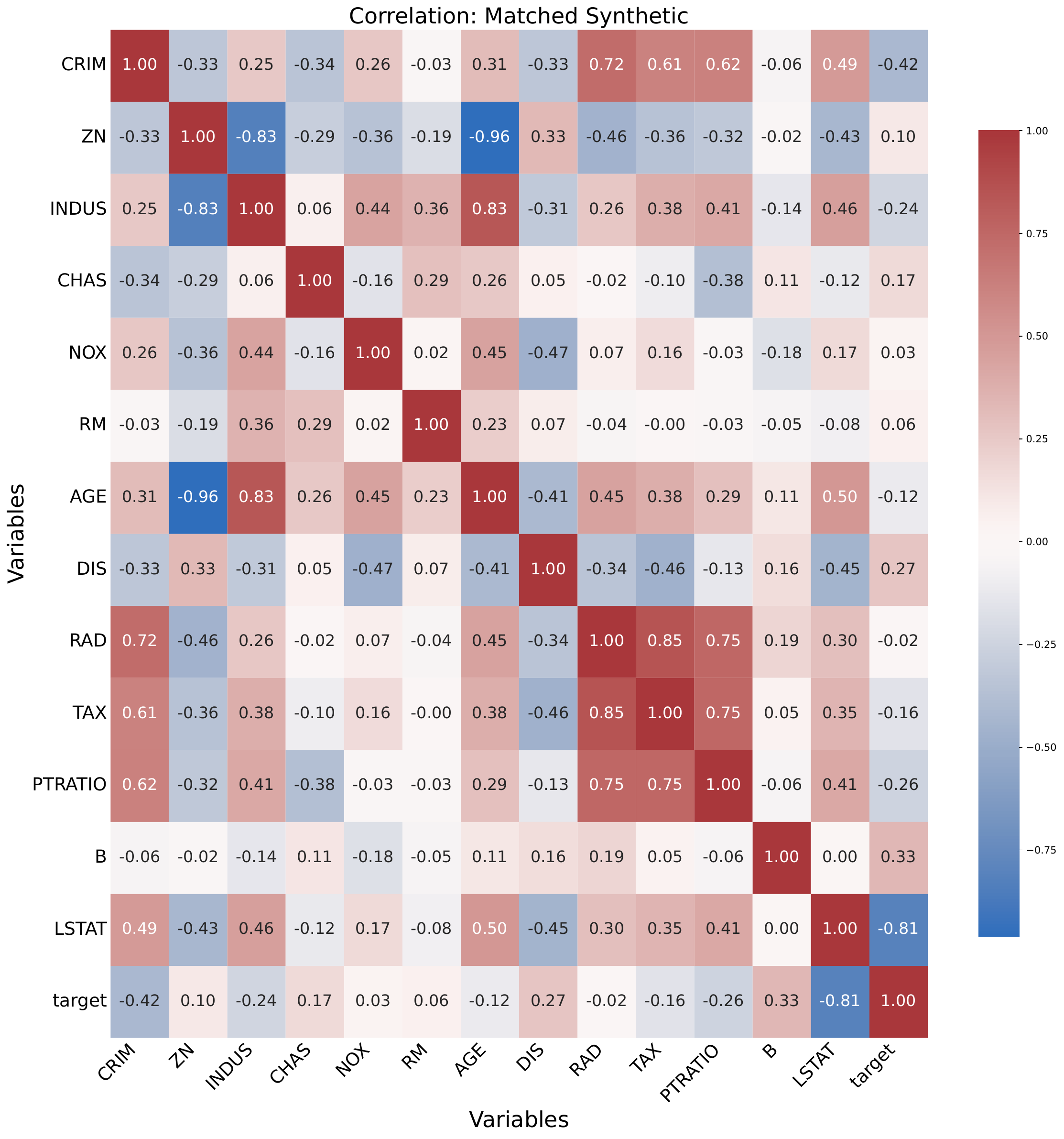}
   \par\smallskip
   {\scriptsize\textbf{(b)} Matched synthetic correlations}
   \label{fig:corr_synth}
 \end{minipage}
 \caption{Comparison of feature correlations for real minority versus GAN-matched synthetic samples. Panel (a) shows the Pearson correlation matrix of the real minority data; panel (b) shows the same for matched synthetic data.}
 \label{fig:corr_comparison}
\end{figure*}

\subsection{t-SNE of real minority vs. matched synthetic samples}

We assess the quality of our nearest-neighbour matching by visualizing real minority and GAN-generated samples in a 2D t-SNE embedding on the Boston dataset. Figure~\ref{fig:tsne_minority_synth} shows embeddings at four perplexity settings. Orange circles mark real minority points, and green diamonds mark their matched synthetic counterparts. Across all panels, the synthetic points closely overlap with real clusters, demonstrating that our matching procedure selects synthetic samples which faithfully follow the true minority distribution at both local (low perplexity) and global (high perplexity) scales.

\subsection{Correlation structure of real and synthetic minority samples}

We compare the pairwise relationships among all 13 features of the Boston dataset (including the target) in the real minority set and in the GAN-matched synthetic set by computing Pearson correlation matrices and visualizing them as heatmaps. Figure~\ref{fig:corr_comparison}(a) shows the correlation matrix for the real minority samples, and Figure~\ref{fig:corr_comparison}(b) displays the same for the matched synthetic samples, which closely mirror the real correlations. To quantify deviations, Figure~\ref{fig:corr_comparison}(c) plots the element-wise difference. Overall, these results demonstrate that our matching procedure preserves the original minority correlation structure across variables.

\begin{figure*}[!t]
  \centering
  \includegraphics[width=\columnwidth]{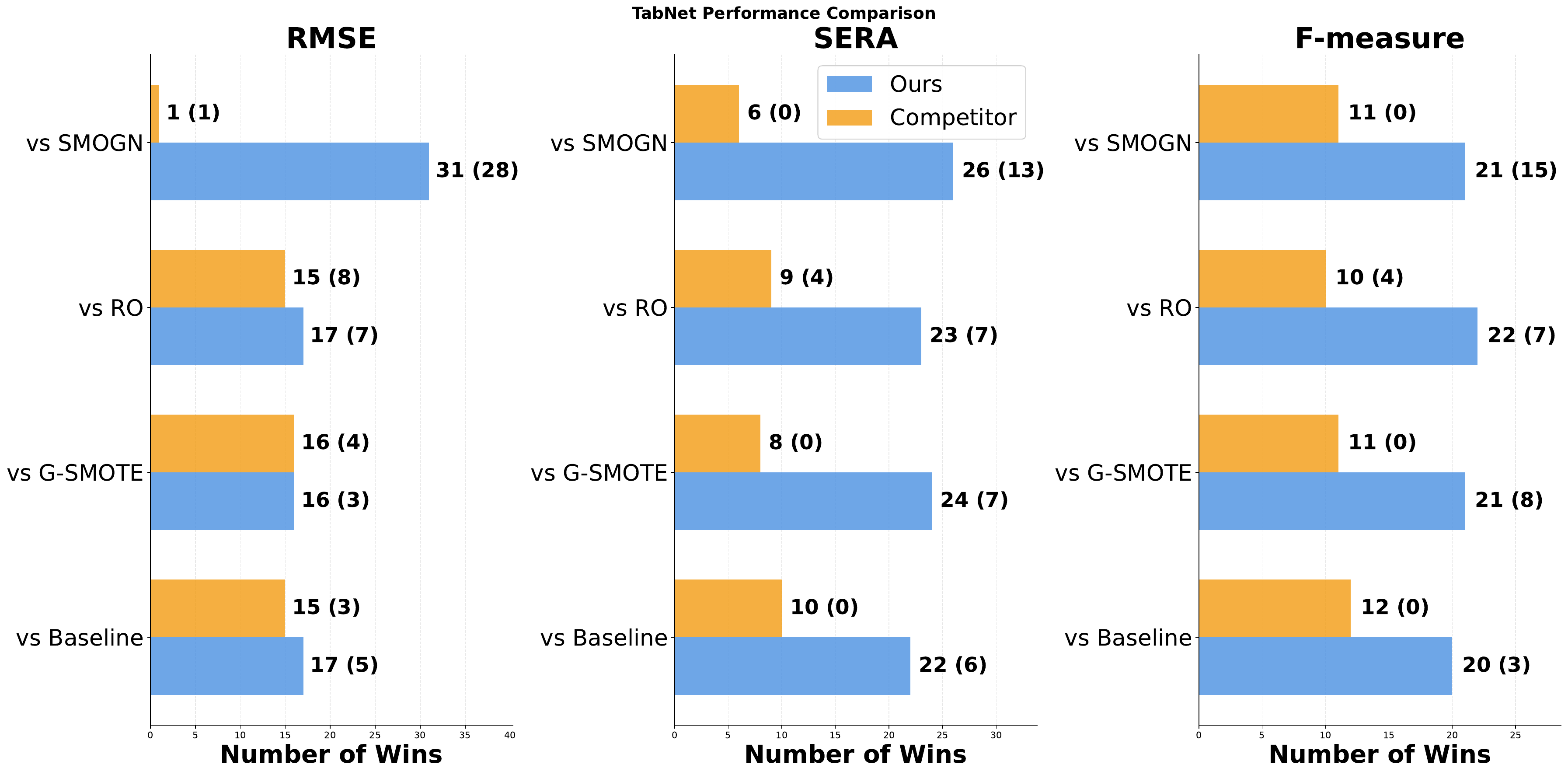}
  \caption{Performance comparison across methods and metrics for TabNet model. Blue bars represent our method's wins, orange bars represent competitor wins. Numbers show total wins with significant wins in parentheses. Higher values indicate better performance.}
  \label{fig:tabnet}
\end{figure*}

\begin{figure*}[!t]
  \centering
  \includegraphics[width=\columnwidth]{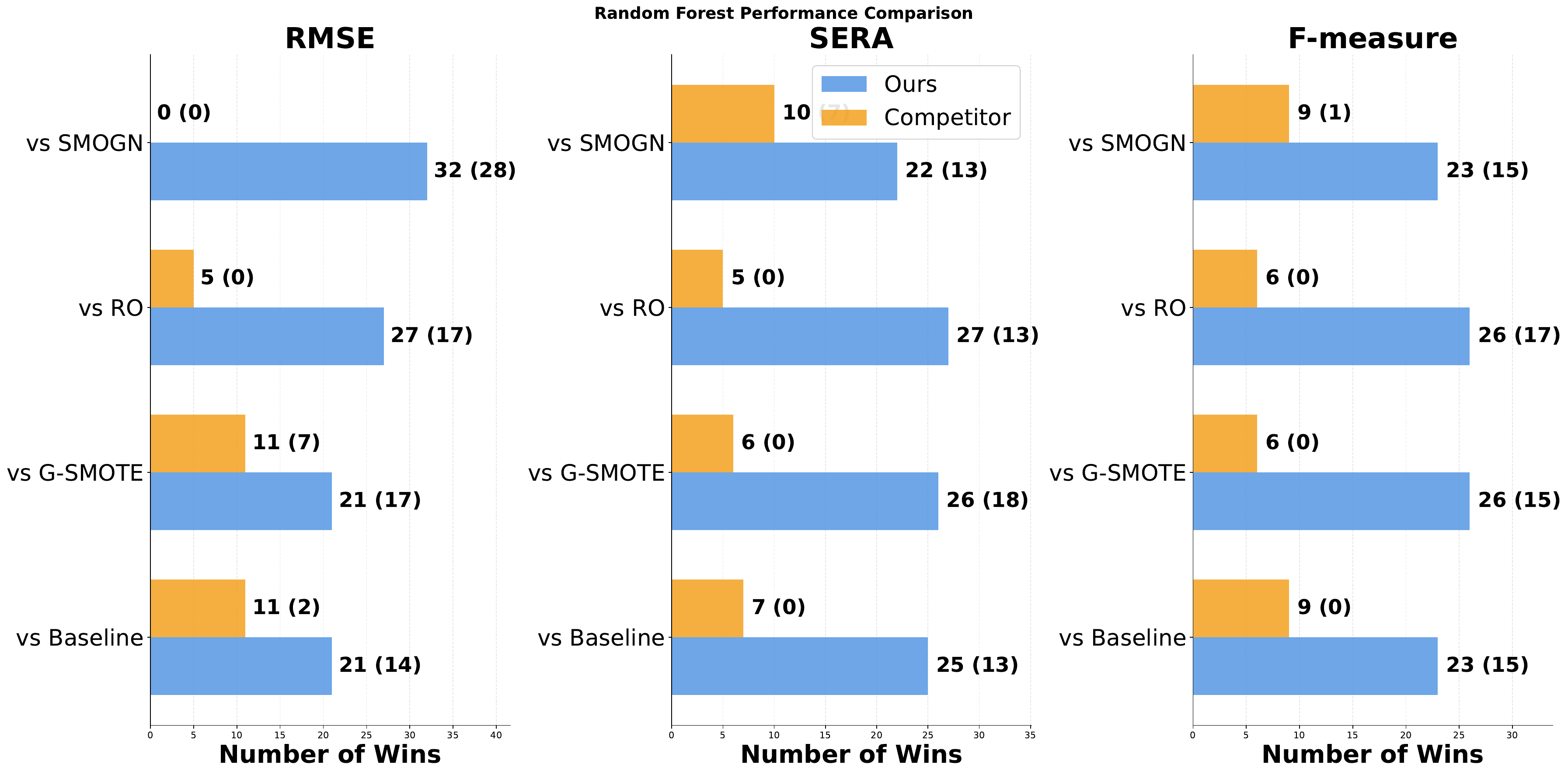}
  \caption{Performance comparison across methods and metrics for random forest model. Blue bars represent our method's wins, orange bars represent competitor wins. Numbers show total wins with significant wins in parentheses. Higher values indicate better performance.}
  \label{fig:rf}
\end{figure*}

\section{Results}

We present the results of comparing our framework with other oversampling methods, including random oversampling (RO), SMOGN, and G-SMOTE, as well as a no-oversampling baseline. In Figures~\ref{fig:tabnet} and \ref{fig:rf}, we report aggregated pairwise comparisons for each evaluation metric (RMSE, SERA, F-measure) for both TabNet and Random Forest models.

For each dataset and each pairwise comparison, a winner is determined based on outcomes across 25 random splits, so each dataset contributes one win per threshold and two wins in total across both thresholds. Specifically, for any two methods under comparison, the method that prevails in the majority of splits is declared the winner. Next, we perform the Wilcoxon signed-rank test for each pairwise comparison. We compute the differences between the 25 split metric values of the two methods to form a difference vector, and then the test assesses whether these differences are statistically significant at an alpha level of 0.05, corresponding to a 5\% risk of incorrectly concluding that a difference exists when there is none \citep{wilcoxon1945}.

In Figures~\ref{fig:tabnet} and \ref{fig:rf}, the numbers outside the parentheses indicate the total number of wins for each method, while the numbers inside the parentheses represent statistically significant wins. Our framework outperforms random oversampling (RO), SMOGN, G-SMOTE and the no-oversampling baseline across all evaluation metrics (RMSE, SERA, F-measure) for both TabNet and Random Forest. It achieves the highest total wins in every pairwise comparison and secures the majority of statistically significant wins, demonstrating that data-driven adversarial and deterministic Mahalanobis matching generate realistic synthetic samples that better align with the true distribution and yield stronger predictive accuracy, even when paired with a powerful regressor like TabNet.

\section{Limitations and Discussion}
Our approach entails hyperparameter choices that may require tuning. In particular, the nearest-neighbour matching parameters ($k$ and $n_{\mathrm{pick}}$) influence the balance between sample diversity and fidelity, and GAN training introduces computational overhead and sensitivity to network architecture and learning rates. Future work will investigate adaptive strategies for selecting matching parameters, more efficient adversarial training schedules, and extensions to high-dimensional or multimodal regression settings. Overall, our results demonstrate that a data-driven, Mahalanobis-GMM-GAN pipeline can substantially improve rare-target prediction, while offering clear directions for practical refinement.

\section{Conclusions}

Our framework delivers a fully data-driven pipeline for augmenting imbalanced regression datasets, removing the need for domain expertise or arbitrary rarity thresholds. Comprehensive statistical tests and qualitative visualizations confirm that our Mahalanobis-GMM detection accurately isolates rare observations in the joint feature-target space, and that the WGAN-GP generation followed by deterministic Mahalanobis matching yields realistic synthetic samples that outperform existing oversampling methods (SMOGN, G-SMOTE, and random oversampling) across all evaluation metrics.

\bibliographystyle{plainnat}
\bibliography{references}

\begin{thebibliography}{56}
\providecommand{\natexlab}[1]{#1}
\providecommand{\url}[1]{\texttt{#1}}
\expandafter\ifx\csname urlstyle\endcsname\relax
  \providecommand{\doi}[1]{doi: #1}\else
  \providecommand{\doi}{doi: \begingroup \urlstyle{rm}\Url}\fi

\bibitem[Aggarwal(2017)]{aggarwal2017introduction}
Charu~C Aggarwal.
\newblock \emph{An introduction to outlier analysis}.
\newblock Springer, 2017.

\bibitem[Alahyari and Domaratzki(2025{\natexlab{a}})]{alahyari2025ldao}
Shayan Alahyari and Mike Domaratzki.
\newblock Local distribution-based adaptive oversampling for imbalanced regression.
\newblock Preprint arXiv:2504.14316, 2025{\natexlab{a}}.

\bibitem[Alahyari and Domaratzki(2025{\natexlab{b}})]{alahyari2025smogan}
Shayan Alahyari and Mike Domaratzki.
\newblock {SMOGAN}: Synthetic minority oversampling with {GAN} refinement for imbalanced regression.
\newblock Preprint arXiv:2504.21152, 2025{\natexlab{b}}.

\bibitem[Alcalá-Fdez et~al.(2011)Alcalá-Fdez, Fernandez, Luengo, Derrac, García, Sánchez, and Herrera]{alcala2011}
J.~Alcalá-Fdez, A.~Fernandez, J.~Luengo, J.~Derrac, S.~García, L.~Sánchez, and F.~Herrera.
\newblock Keel data-mining software tool: Data set repository, integration of algorithms and experimental analysis framework.
\newblock \emph{Journal of Multiple-Valued Logic and Soft Computing}, 17\penalty0 (2-3):\penalty0 255--287, 2011.

\bibitem[Ar{\i}k and Pfister(2021)]{arik2021tabnet}
Sercan~{\"O}. Ar{\i}k and Tomas Pfister.
\newblock Tabnet: Attentive interpretable tabular learning.
\newblock In \emph{Proceedings of the AAAI Conference on Artificial Intelligence}, volume~35, pages 6679--6687, 2021.
\newblock \doi{10.1609/aaai.v35i8.16826}.

\bibitem[Arjovsky et~al.(2017)Arjovsky, Chintala, and Bottou]{arjovsky2017wasserstein}
Martin Arjovsky, Soumith Chintala, and Léon Bottou.
\newblock Wasserstein {GAN}.
\newblock \emph{arXiv preprint arXiv:1701.07875v3 [stat.ML]}, 2017.

\bibitem[Balakrishnan and Shenoy(2023)]{balakrishnan2023mmote}
Kavitha Balakrishnan and Anitha Shenoy.
\newblock Mmote: A {Mahalanobis} distance-based oversampling approach for improving minority class classification in parkinson’s disease detection.
\newblock \emph{Applied Soft Computing}, 136:\penalty0 110174, 2023.
\newblock \doi{10.1016/j.asoc.2023.110174}.

\bibitem[Branco et~al.(2016)Branco, Torgo, and Ribeiro]{branco2016}
P.~Branco, L.~Torgo, and R.~P. Ribeiro.
\newblock A survey of predictive modeling under imbalanced distributions.
\newblock \emph{ACM Computing Surveys}, 49\penalty0 (2):\penalty0 Article 31, 2016.

\bibitem[Branco et~al.(2017)Branco, Torgo, and Ribeiro]{branco2017}
P.~Branco, L.~Torgo, and R.~P. Ribeiro.
\newblock {SMOGN}: A pre-processing approach for imbalanced regression.
\newblock In \emph{Proceedings of Machine Learning Research: LIDTA}, volume~74, pages 36--50, 2017.

\bibitem[Branco et~al.(2019)Branco, Torgo, and Ribeiro]{branco2019}
P.~Branco, L.~Torgo, and R.~P. Ribeiro.
\newblock Pre-processing approaches for imbalanced distributions in regression.
\newblock \emph{Neurocomputing}, 343:\penalty0 76--99, 2019.

\bibitem[Breiman(2001)]{breiman2001random}
Leo Breiman.
\newblock Random forests.
\newblock \emph{Machine Learning}, 45\penalty0 (1):\penalty0 5--32, 2001.
\newblock \doi{10.1023/A:1010933404324}.

\bibitem[Buda et~al.(2018)Buda, Maki, and Mazurowski]{buda2018}
M.~Buda, A.~Maki, and M.~A. Mazurowski.
\newblock A systematic study of the class imbalance problem in convolutional neural networks.
\newblock \emph{Neural Networks}, 106:\penalty0 249--259, 2018.

\bibitem[Camacho and Bacao(2024)]{camacho2024}
L.~Camacho and F.~Bacao.
\newblock {WSMOTER}: A novel approach for imbalanced regression.
\newblock \emph{Applied Intelligence}, 54:\penalty0 8789--8799, 2024.

\bibitem[Camacho et~al.(2022)Camacho, Douzas, and Bacao]{camacho2022}
L.~Camacho, G.~Douzas, and F.~Bacao.
\newblock Geometric {SMOTE} for regression.
\newblock \emph{Expert Systems with Applications}, 193:\penalty0 116387, 2022.

\bibitem[Chawla et~al.(2002)Chawla, Bowyer, Hall, and Kegelmeyer]{chawla2002}
N.~V. Chawla, K.~W. Bowyer, L.~O. Hall, and W.~P. Kegelmeyer.
\newblock {SMOTE}: Synthetic minority over-sampling technique.
\newblock \emph{Journal of Artificial Intelligence Research}, 16:\penalty0 321--357, 2002.

\bibitem[Chawla et~al.(2004)Chawla, Japkowicz, and Kolcz]{chawla2004}
N.~V. Chawla, N.~Japkowicz, and A.~Kolcz.
\newblock Editorial: Special issue on learning from imbalanced data sets.
\newblock \emph{ACM SIGKDD Explorations Newsletter}, 6\penalty0 (1):\penalty0 1--6, 2004.

\bibitem[Domingos(1999)]{domingos1999}
P.~Domingos.
\newblock Metacost: A general method for making classifiers cost-sensitive.
\newblock In \emph{Proceedings of the 5th ACM SIGKDD International Conference on Knowledge Discovery and Data Mining (KDD)}, pages 155--164, 1999.

\bibitem[Elkan(2001)]{elkan2001}
C.~Elkan.
\newblock The foundations of cost-sensitive learning.
\newblock In \emph{Proceedings of the 17th International Joint Conference on Artificial Intelligence (IJCAI)}, pages 973--978, 2001.

\bibitem[Engelmann and Lessmann(2020)]{engelmann2020conditional}
Justin Engelmann and Stefan Lessmann.
\newblock Conditional {Wasserstein} {GAN}‐based oversampling of tabular data for imbalanced learning.
\newblock Preprint arXiv:2008.09202v1, 2020.

\bibitem[Ghorbani(2019)]{ghorbani2019mahalanobis}
Hamid Ghorbani.
\newblock Mahalanobis distance and its application for detecting multivariate outliers.
\newblock \emph{Facta Universitatis, Series: Mathematics and Informatics}, pages 583--595, 2019.

\bibitem[Goodfellow et~al.(2016)Goodfellow, Bengio, and Courville]{Goodfellow-et-al-2016}
Ian Goodfellow, Yoshua Bengio, and Aaron Courville.
\newblock \emph{Deep Learning}.
\newblock MIT Press, 2016.
\newblock \url{http://www.deeplearningbook.org}.

\bibitem[Gulrajani et~al.(2017)Gulrajani, Ahmed, Arjovsky, Dumoulin, and Courville]{gulrajani2017improved}
Ishaan Gulrajani, Faruk Ahmed, Martin Arjovsky, Vincent Dumoulin, and Aaron Courville.
\newblock Improved training of {Wasserstein} {GANs}.
\newblock In \emph{Advances in Neural Information Processing Systems 30 (NeurIPS 2017)}, pages 5767--5777, 2017.

\bibitem[Guo et~al.(2017)Guo, Li, Shang, Gu, Huang, and Gong]{haixiang2017}
H.~Guo, Y.~Li, J.~Shang, M.~Gu, Y.~Huang, and B.~Gong.
\newblock Learning from class-imbalanced data: Review of methods and applications.
\newblock \emph{Expert Systems with Applications}, 73:\penalty0 220--239, 2017.

\bibitem[He and Garcia(2009)]{he2009}
H.~He and E.~A. Garcia.
\newblock Learning from imbalanced data.
\newblock \emph{IEEE Transactions on Knowledge and Data Engineering}, 21\penalty0 (9):\penalty0 1263--1284, 2009.

\bibitem[Jiang et~al.(2019)Jiang, Hong, Zhou, He, and Cheng]{jiang2019gan}
Wenqian Jiang, Yang Hong, Beitong Zhou, Xin He, and Cheng Cheng.
\newblock A {GAN}‐based anomaly detection approach for imbalanced industrial time series.
\newblock \emph{IEEE Access}, 7:\penalty0 143608--143619, 2019.

\bibitem[Johnson and Khoshgoftaar(2019)]{johnson2019}
J.~M. Johnson and T.~M. Khoshgoftaar.
\newblock Survey on deep learning with class imbalance.
\newblock \emph{Journal of Big Data}, 6\penalty0 (1):\penalty0 1--54, 2019.

\bibitem[Kamangir et~al.(2024)Kamangir, Sams, Dokoozlian, Sanchez, and Earles]{kamangir2024large}
Hossein Kamangir, S.~Sams, B.\, Nuno Dokoozlian, Luis Sanchez, and M.~Earles, J.\.
\newblock Large-scale spatio-temporal yield estimation via deep learning using satellite and management data fusion in vineyards.
\newblock \emph{Computers and Electronics in Agriculture}, 216:\penalty0 108439, 2024.

\bibitem[Krawczyk(2016)]{krawczyk2016}
B.~Krawczyk.
\newblock Learning from imbalanced data: open challenges and future directions.
\newblock \emph{Progress in Artificial Intelligence}, 5\penalty0 (4):\penalty0 221--232, 2016.

\bibitem[Kunz(2020)]{kunz2020}
N.~Kunz.
\newblock Smogn: Synthetic minority over-sampling technique for regression with gaussian noise.
\newblock PyPI, version v0.1.2, 2020.

\bibitem[Lee and Park(2021)]{lee2019ganids}
JooHwa Lee and KeeHyun Park.
\newblock Gan-based imbalanced data intrusion detection system.
\newblock \emph{Personal and Ubiquitous Computing}, 25\penalty0 (1):\penalty0 121--128, 2021.

\bibitem[Liu et~al.(2009)Liu, Wu, and Zhou]{liu2009}
X.-Y. Liu, J.~Wu, and Z.-H. Zhou.
\newblock Exploratory undersampling for class-imbalance learning.
\newblock \emph{IEEE Transactions on Systems, Man, and Cybernetics, Part B (Cybernetics)}, 39\penalty0 (2):\penalty0 539--550, 2009.

\bibitem[Ma et~al.(2024)Ma, Huang, Nan, Moniz, Zhang, Wiest, and Chawla]{ma2024revisiting}
Yihong Ma, Xiaobao Huang, Bozhao Nan, Nuno Moniz, Xiangliang Zhang, Olaf Wiest, and Nitesh~V. Chawla.
\newblock Are we making much progress? revisiting chemical reaction yield prediction from an imbalanced regression perspective.
\newblock In \emph{Companion Proceedings of the ACM Web Conference 2024 (WWW '24 Companion)}, pages 791--794. ACM, 2024.

\bibitem[Manly(2005)]{manly2005multivariate}
Bryan F.~J. Manly.
\newblock \emph{Multivariate Statistical Methods: A Primer}.
\newblock Chapman \& Hall/CRC Press, Boca Raton, FL ; London, 3rd edition, 2005.

\bibitem[Mariani et~al.(2018)Mariani, Scheidegger, Istrate, Bekas, and Malossi]{mariani2018bagan}
Giovanni Mariani, Florian Scheidegger, Roxana Istrate, Costas Bekas, and Cristiano Malossi.
\newblock {BAGAN}: Data augmentation with balancing {GAN}.
\newblock Preprint arXiv:1803.09655v2, 2018.

\bibitem[McLachlan and Krishnan(2008)]{mclachlan2008algorithm}
Geoffrey~J McLachlan and Thriyambakam Krishnan.
\newblock \emph{The EM algorithm and extensions}.
\newblock John Wiley \& Sons, 2008.

\bibitem[Moniz et~al.(2018)Moniz, Torgo, and Soares]{moniz2018}
N.~Moniz, L.~Torgo, and C.~Soares.
\newblock {SMOTEBoost} for regression: Improving the prediction of extreme values.
\newblock In \emph{Proceedings of the 5th International Conference on Data Science and Advanced Analytics (DSAA)}, pages 127--136, 2018.

\bibitem[NezhadShokouhi et~al.(2020)NezhadShokouhi, Majidi, and Rasoolzadegan]{nezhadshokouhi2020}
Mohammad~Mahdi NezhadShokouhi, Mohammad~Ali Majidi, and Abbas Rasoolzadegan.
\newblock Software defect prediction using over-sampling and feature extraction based on mahalanobis distance.
\newblock \emph{The Journal of Supercomputing}, 76\penalty0 (1):\penalty0 602--635, 2020.
\newblock \doi{10.1007/s11227-019-03051-w}.

\bibitem[Ren et~al.(2022)Ren, Luo, and Urtasun]{ren2022}
M.~Ren, W.~Luo, and R.~Urtasun.
\newblock Balanced mse for imbalanced visual regression.
\newblock In \emph{Proceedings of the IEEE/CVF Conference on Computer Vision and Pattern Recognition (CVPR)}, pages 418--427, 2022.

\bibitem[Ribeiro and Moniz(2020)]{ribeiro2020}
R.~P. Ribeiro and N.~Moniz.
\newblock Imbalanced regression and extreme value prediction.
\newblock \emph{Machine Learning}, 109\penalty0 (9-10):\penalty0 1803--1835, 2020.

\bibitem[Ribeiro(2011)]{ribeiro2011a}
R.~P.~A. Ribeiro.
\newblock \emph{Utility-based regression}.
\newblock PhD thesis, Faculty of Sciences, University of Porto, Porto, 2011.

\bibitem[Ribeiro~Junior et~al.(2023)Ribeiro~Junior, de~Almeida, Jorge, Pereira, Francisco, and Gomes]{ribeiro2023gmmmaha}
Ronny~Francis Ribeiro~Junior, Fabricio~Alves de~Almeida, Ariosto~Bretanha Jorge, João Luiz~Junho Pereira, Matheus~Brendon Francisco, and Guilherme~Ferreira Gomes.
\newblock On the use of the {Gaussian} mixture model and the {Mahalanobis} distance for fault diagnosis in dynamic components of electric motors.
\newblock \emph{Journal of the Brazilian Society of Mechanical Sciences and Engineering}, 45:\penalty0 139, 2023.
\newblock \doi{10.1007/s40430-023-04056-6}.

\bibitem[Scheepens et~al.(2023)Scheepens, Müller, Vos, Giannakas, Portilla-Figueroa, Prieto, and Correoso]{scheepens2023adapting}
R.~Scheepens, D.\, R.~Müller, J.~Vos, I.~Giannakas, E.~Portilla-Figueroa, L.~Prieto, and F.~Correoso.
\newblock Adapting a deep convolutional recurrent neural network model for improved spatio-temporal forecasting of extreme wind speed events using imbalanced regression losses.
\newblock \emph{Geoscientific Model Development}, 16:\penalty0 251--270, 2023.

\bibitem[Sharma et~al.(2022)Sharma, Singh, and Chandra]{sharma2022smotified}
Anuraganand Sharma, Prabhat~Kumar Singh, and Rohitash Chandra.
\newblock {SMOTified}-{GAN} for class imbalanced pattern classification problems.
\newblock \emph{IEEE Access}, 10:\penalty0 30655--30665, 2022.

\bibitem[Steininger et~al.(2021)Steininger, Kobs, Davidson, Krause, and Hotho]{steininger2021}
M.~Steininger, K.~Kobs, P.~Davidson, A.~Krause, and A.~Hotho.
\newblock Density-based weighting for imbalanced regression.
\newblock \emph{Machine Learning}, 110\penalty0 (8):\penalty0 2187--2210, 2021.

\bibitem[Tanaka and Aranha(2019)]{tanaka2019dataaugmentation}
Fabio Henrique Kiyoiti dos~Santos Tanaka and Claus Aranha.
\newblock Data augmentation using {GANs}.
\newblock In \emph{Proceedings of Machine Learning Research}, volume XXX, pages 1--16, 2019.

\bibitem[Torgo et~al.(2013)Torgo, Ribeiro, da~Costa, and Pal]{torgo2013}
L.~Torgo, R.~P. Ribeiro, J.~P. da~Costa, and S.~Pal.
\newblock {SMOTE} for regression.
\newblock In \emph{Intelligent Data Engineering and Automated Learning (IDEAL 2013). Lecture Notes in Computer Science}, volume 8206, pages 378--387, 2013.

\bibitem[Torgo and Ribeiro(2007)]{torgo2007utility}
Luis Torgo and Rita Ribeiro.
\newblock Utility-based regression.
\newblock In \emph{Proceedings of the 11th European Conference on Principles and Practice of Knowledge Discovery in Databases (PKDD 2007)}, pages 597--604, 2007.

\bibitem[Torgo and Ribeiro(2009)]{torgo2009}
Luis Torgo and Rita Ribeiro.
\newblock Precision and recall for regression.
\newblock In \emph{Discovery Science (DS 2009)}, volume 5808 of \emph{Lecture Notes in Artificial Intelligence}, pages 332--346. Springer‑Verlag Berlin Heidelberg, 2009.

\bibitem[van~der Maaten and Hinton(2008)]{vandermaaten2008visualizing}
Laurens van~der Maaten and Geoffrey Hinton.
\newblock Visualizing data using t-sne.
\newblock \emph{Journal of Machine Learning Research}, 9:\penalty0 2579--2605, 2008.

\bibitem[Ververidis and Kotropoulos(2008)]{ververidis2008}
Dimitrios Ververidis and Constantine Kotropoulos.
\newblock Gaussian mixture modeling by exploiting the mahalanobis distance.
\newblock \emph{IEEE Transactions on Signal Processing}, 56\penalty0 (7):\penalty0 2797--2807, July 2008.
\newblock \doi{10.1109/TSP.2008.917350}.

\bibitem[Wilcoxon(1945)]{wilcoxon1945}
F.~Wilcoxon.
\newblock Individual comparisons by ranking methods.
\newblock \emph{Biometrics Bulletin}, 1\penalty0 (6):\penalty0 80--83, 1945.

\bibitem[Wu et~al.(2022)Wu, Kunz, and Branco]{wu2022}
W.~Wu, N.~Kunz, and P.~Branco.
\newblock Imbalancedlearningregression-a python package to tackle the imbalanced regression problem.
\newblock In \emph{Joint European Conference on Machine Learning and Knowledge Discovery in Databases}, pages 645--648, 2022.

\bibitem[Yang et~al.(2021)Yang, Xie, Yu, He, and Liu]{yang2021}
J.~Yang, L.~Xie, Q.~Yu, X.~He, and J.~Liu.
\newblock Delving into deep imbalanced regression.
\newblock In \emph{Proceedings of the 38th International Conference on Machine Learning (ICML)}, pages 8437--8447, 2021.

\bibitem[Yao and Lin(2021)]{yao2021emdo}
Leehter Yao and Tung-Bin Lin.
\newblock Evolutionary mahalanobis distance-based oversampling for multi-class imbalanced data classification.
\newblock \emph{Sensors}, 21\penalty0 (19):\penalty0 6616, 2021.
\newblock \doi{10.3390/s21196616}.

\bibitem[Zhang et~al.(2025)Zhang, Wang, Chen, Li, Phan, and Chench]{zhang2025irmae}
Qian Zhang, Yue Wang, Li~Chen, Hao Li, Thanh~N. Phan, and Peng Chench.
\newblock Irmae-akde: A novel deep imbalanced regression approach for performance prediction of rolled steel plate yield strength.
\newblock \emph{IEEE Transactions on Instrumentation and Measurement}, 74:\penalty0 2006911, 2025.

\bibitem[Zhou and Liu(2010)]{zhou2010}
Z.-H. Zhou and X.-Y. Liu.
\newblock On multi-class cost-sensitive learning.
\newblock \emph{Computational Intelligence}, 26\penalty0 (3):\penalty0 232--257, 2010.

\end{thebibliography}

\end{document}